\documentclass[conference]{IEEEtran}
\IEEEoverridecommandlockouts
\usepackage[maxbibnames=5, minbibnames=5, natbib=true, maxcitenames=2, mincitenames=1, style=ieee, doi=false, isbn=false, url=false, eprint=false]{biblatex}
\usepackage{amsmath,amssymb,amsfonts}
\usepackage{algorithmic}
\usepackage{graphicx}
\usepackage{textcomp}
\usepackage{xcolor}
\usepackage[justification=raggedright, position=top, ,singlelinecheck=false]{subcaption}
\usepackage[hidelinks]{hyperref}
\usepackage{booktabs}
\usepackage{xurl}
\usepackage[final]{microtype}

\def\BibTeX{{\rm B\kern-.05em{\sc i\kern-.025em b}\kern-.08em
    T\kern-.1667em\lower.7ex\hbox{E}\kern-.125emX}}

\definecolor{myblue}{RGB}{39, 111, 176}
\definecolor{myred}{RGB}{186, 40, 50}
\definecolor{myviolet}{RGB}{140, 103, 175}
\definecolor{myred2}{RGB}{216, 27, 96}
\definecolor{myblue2}{RGB}{30, 136, 229}
\definecolor{myred3}{RGB}{255, 102, 102}
\definecolor{myblue3}{RGB}{141, 165, 237}

\DeclareRobustCommand{\IEEEauthorrefmark}[1]{\smash{\textsuperscript{\footnotesize #1}}}
\newcommand{\authorcomma}{\smash{\textsuperscript{\footnotesize ,}}}

\begin{document}

\title{Towards Physically Consistent Deep Learning For Climate Model Parameterizations
}

\author{
\IEEEauthorblockN{
Birgit K{\"u}hbacher\IEEEauthorrefmark{1}\authorcomma\IEEEauthorrefmark{2}\authorcomma\IEEEauthorrefmark{3}, 
Fernando Iglesias-Suarez\IEEEauthorrefmark{1},  
Niki Kilbertus\IEEEauthorrefmark{2}\authorcomma\IEEEauthorrefmark{3}\authorcomma\IEEEauthorrefmark{4}, 
Veronika Eyring\IEEEauthorrefmark{1}\authorcomma\IEEEauthorrefmark{2}\authorcomma\IEEEauthorrefmark{5}
}

\IEEEauthorblockA{\IEEEauthorrefmark{1}Deutsches Zentrum f\"ur Luft- und Raumfahrt (DLR), Institut f\"ur Physik der Atmosph\"are, Oberpfaffenhofen, Germany}
\IEEEauthorblockA{\IEEEauthorrefmark{2}Technical University of Munich, Munich, Germany}
\IEEEauthorblockA{\IEEEauthorrefmark{3}Helmholtz Munich, Munich, Germany}
\IEEEauthorblockA{\IEEEauthorrefmark{4}Munich Center for Machine Learning (MCML), Munich, Germany}
\IEEEauthorblockA{\IEEEauthorrefmark{5}University of Bremen, Institute of Environmental Physics, Bremen, Germany}
\IEEEauthorblockA{Corresponding author: birgit.kuehbacher@dlr.de}
}

\maketitle


\begin{abstract}
Climate models play a critical role in understanding and projecting climate change. 
Due to their complexity, their horizontal resolution of about 40-100 km remains too coarse to resolve processes such as clouds and convection, which need to be approximated via parameterizations. 
These parameterizations are a major source of systematic errors and large uncertainties in climate projections. 
Deep learning (DL)-based parameterizations, trained on data from computationally expensive short, high-resolution simulations, have shown great promise for improving climate models in that regard. 
However, their lack of interpretability and tendency to learn spurious non-physical correlations result in reduced trust in the climate simulation. 
We propose an efficient supervised learning framework for DL-based parameterizations that leads to physically consistent models with improved interpretability and negligible computational overhead compared to standard supervised training.
First, key features determining the target physical processes are uncovered.
Subsequently, the neural network is fine-tuned using only those relevant features. 
We show empirically that our method robustly identifies a small subset of the inputs as actual physical drivers, therefore removing spurious non-physical relationships. 
This results in by design physically consistent and interpretable neural networks while maintaining the predictive performance of unconstrained black-box DL-based parameterizations. 
\end{abstract}

\begin{IEEEkeywords} 
climate modeling, physical consistency, deep learning, subgrid parameterization, interpretability
\end{IEEEkeywords}


\section{Introduction}
\label{sec:intro}

Impacts of climate change, such as wildfires, droughts, and loss of biodiversity, threaten both human well-being and the health of the planet \citep{ipcc2023}. 
Climate models are crucial in understanding these changes and for providing climate projections that deliver important information for mitigation and adaptation strategies \citep{ipcc2023, ipccAR623synthesisSPM}. 
Climate models project climate change over several decades to hundreds of years for a variety of plausible future scenarios \citep{tebaldi2021a}. 
However, due to their complexity, the models' horizontal grid resolution in the atmosphere remains coarse ($\sim$40 to 100 kilometers \citep{ANNEXII2021ipcc}). 
This resolution is too coarse to explicitly simulate convective and other important small-scale processes.
For instance, cloud formation takes place at scales ranging from 10 to 100 meters \citep{schneider2017}, yet these processes play a pivotal role in the climate system. 
Clouds transport heat and moisture and have a large impact on radiation, either by reflecting or absorbing it \citep{Boucher2013clouds}. 
Thus, such unresolved subgrid-scale processes need to be parameterized in climate models \citep{gentine2021}, which forms a major source of long-standing systematic errors \citep{eyring2021ipcc} and uncertainties in climate projections \citep{tebaldi2021a}. 
High-resolution km-scale climate models can alleviate a number of these biases \citep{stevens2019}, but due to their computational costs, they can currently not provide climate projections for multiple decades or longer \citep{gentine2021}. 

The development of hybrid models presents a promising approach for long-term climate projections. 
These models improve subgrid-scale parameterizations with machine learning, particularly deep learning (DL) \citep{gentine2021,eyring2024a}, and are efficient enough to generate large ensembles, which are crucial for simulating internal variability and extreme events.
In such hybrid models, climate models are coupled with DL parameterizations trained on data from short, high-resolution climate simulations. 
The resulting simulations show reduced systematic errors compared to the host climate model using the traditional parameterization \citep{rasp2018, gentine2021, grundner2022, iglesias-suarez2024} at higher computational efficiency than high-resolution simulations \citep{gentine2018a, henn2024}. 
However, the black-box nature of DL models and their tendency to learn spurious non-physical correlations poses challenges in understanding their prediction-making processes \citep{gilpin2018, zhang2021a} and in providing out-of-distribution climate projections, leading to reduced confidence in neural network (NN) predictions. 
Model interpretability is especially important in Earth system sciences, where models should be consistent with our physical knowledge \citep{kashinath2021, toms2020}. 
Furthermore, there is strong interest in utilizing these models not only for prediction but also to enhance our understanding of the physical systems under investigation \citep{camps-valls2020, jiang2024a}. 

In this work, we introduce the Physically Consistent Masking (PCMasking) framework, developed specifically to build predictive models that are, by design, physically consistent and interpretable.\footnote{Code available at \url{https://github.com/EyringMLClimateGroup/kuehbacher24ICMLA_PCMasking}.} 
During the initial phase of an automated training procedure, NNs in our PCMasking framework implicitly uncover key physical input features while learning the climate model parameterization. 
Subsequently, training focuses on fine-tuning model weights using only physically consistent input features. 
The PCMasking framework is distinguished by two primary attributes:  
(1) The internal architecture of the NNs can be customized to suit the specific task. 
This adaptability positions the PCMasking framework as a versatile extension for facilitating physical driver selection not only in climate model parameterizations but also in other applications.
(2) Unlike other approaches to DL-based climate model parameterization (e.g., \citep{bolton2019, beucler2020, guan2023}), the PCMasking framework is purely data-driven, requiring no prior information about physical mechanisms.
This is an advantage, as subgrid-scale processes are complex and not fully understood. 
\citet{iglesias-suarez2024} developed a causally-informed DL approach that also achieves physical consistency without explicitly incorporating physical constraints
However, this method comes with the caveat of a computationally expensive causal discovery process, which requires extensive domain knowledge.
In contrast, PCMasking is a coherent and mostly automated framework, unique in its efficiency and usability without compromising performance in terms of prediction compared to existing techniques. 

Sec.~\ref{sec:related_work} provides a brief summary of related work on DL-based subgrid parameterizations. 
In Sec.~\ref{sec:method}, we introduce the PCMasking framework and evaluate its offline performance using data from the Superparameterized Community Atmosphere Model v3.0 (SPCAM) \citep{collins2006a} in Sec.~\ref{sec:experiments}. 
We conclude by discussing limitations and directions for future work.


\section{Related Work}
\label{sec:related_work}

Neural networks offer a promising approach for replacing subgrid-scale physical processes in coarse-resolution climate models as they are able to learn arbitrary nonlinear functions. 
Methods can be broadly distinguished by examining the type of neural network, the parameterization that is to be replaced, and the kind of data used for training.  
\citet{gentine2018a} and \citet{rasp2018} use a feed-forward NN to replace the subgrid-scale convection parameterization and radiation scheme in the Superparameterized Community Atmosphere Model v3.0 (SPCAM) \citep{collins2006a} in an aquaplanet setup.
\citet{grundner2022, henn2024} present work on cloud parameterizations by training feed-forward NNs on coarse-grained high-resolution data.  
\citet{wang2022b} work on a convection parameterization that uses data from multiple atmospheric model columns to improve the offline prediction of their trained NN. 
While previous methods primarily employ feed-forward neural networks, making them deterministic in nature, there are also examples of stochastic parameterization approaches using neural networks. 
These include generative models like generative adversarial networks \cite{perezhogin2023} and variational autoencoders \cite{behrens2022, behrens2024}.

Although these DL-based approaches for climate model parameterizations demonstrate promising performance to varying degrees, they all suffer from a lack of interpretability and physical consistency. 
While some works use interpretability techniques retrospectively to understand network behavior \citep{grundner2022, behrens2022, wang2022b}, the primary focus remains on performance. 
Meanwhile, physical consistency is passively tested for but not built into the DL models.
\citet{brenowitz2020a} highlight this lack of a priori interpretability as one of the limitations of machine learning-based parameterizations. 
However, there are several examples of studies focusing on physical consistency. 
\citet{beucler2020} developed a physically consistent convection parameterization using a feed-forward NN by adapting the loss function and the architecture or rescaling the data. 
Both \citet{bolton2019} and \citet{guan2023} similarly incorporate physical constraints when training convolutional neural networks for subgrid-scale parameterizations. 
\citet{iglesias-suarez2024} aim for physical consistency and better interpretability, opting for a causal discovery method rather than incorporating physical knowledge into their NN.  

While these approaches represent progress in creating DL-based climate model parameterizations that are both physically consistent and interpretable, they rely heavily on either correctly integrating physical knowledge or potentially costly causal discovery in a pre-step. 
In this paper, we extend previous work with a neural network parameterization framework that is both physically consistent and interpretable by design. 


\section{PCMasking Framework}
\label{sec:method}

\begin{figure}[tb]
\centering
\includegraphics[width=\linewidth]{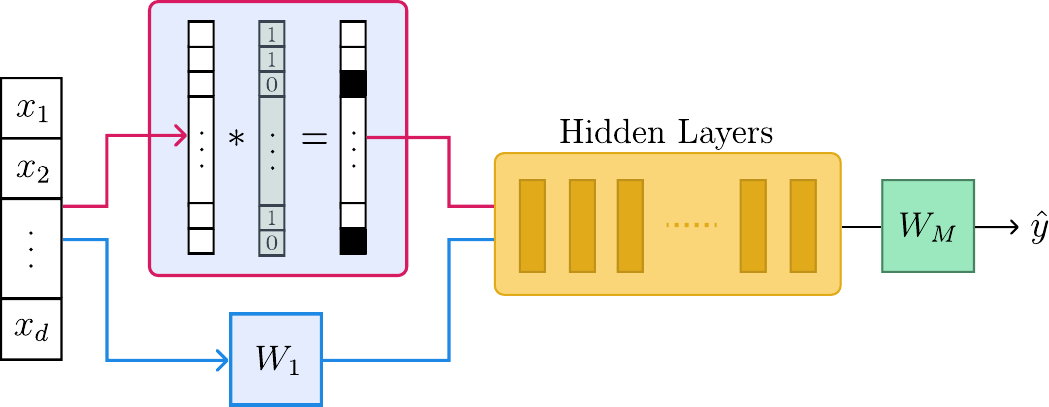}
\caption{Schematic of a neural network model in the PCMasking framework. 
The {\color{myblue2}blue} line indicates the path of the input vector in the pre-masking phase, where it passes through a conventional dense input layer with weight matrix $W_1$. 
In the masking phase, as illustrated by the {\color{myred2}red} line, the input vector is element-wise multiplied with a masking vector. 
In both cases, the information then flows through an arbitrary network architecture before reaching the linear output layer with weight matrix $W_M$. }
\label{fig:pcm_model}
\vspace{-2mm}
\end{figure}

The PCMasking framework is designed to train physically consistent neural networks for climate model parameterizations in a supervised setting. 
First, during the \emph{pre-masking phase}, the input vector passes through a conventional dense input layer, then moves through an arbitrary network architecture, and ultimately reaches the output layer. 
This process is depicted by the blue path in Fig.~\ref{fig:pcm_model}.  
After a certain number of epochs, processing of the input vector is altered, as shown by the red path in Fig.~\ref{fig:pcm_model}. 
In this subsequent \emph{masking phase}, certain input features are masked out by element-wise multiplication with a binary vector, with the goal of only using physically relevant inputs for the prediction. 
We now describe the pre-masking and masking phases in detail.

\subsection{Initial Training and Finding Physical Relationships}

Training in the pre-masking phase serves two purposes. 
The first goal is to predict the value of the output variable from the input features via supervised training. 
Second, we aim to implicitly learn the physical relationships between the input features and the output variable. 
The flow of information in this stage is depicted by the blue path in Fig.~\ref{fig:pcm_model}, and we describe the network as operating in pre-mask mode.

The loss in pre-mask mode is comprised of a prediction loss in the form of the mean squared error and a weighted L1-regularization to encourage sparsity in the input layer weight matrix. 
While regularization is a common technique to prevent overfitting and increase robustness, we use sparsity regularization in particular, as we expect only a limited number of the inputs to be actual physical drivers of the output. 
The regularization term is computed as the entry-wise L1-norm, denoted by $|| \, vec(.) \, ||_1$, applied to the input layer kernel $W_1$. 
This kernel is a $(d \times d)$-dimensional matrix, where $d$ is the number of input features.
In order to make the regularization term independent of the dimensions of $W_1$, and therefore of the number of input features $x_i$, the L1-norm is scaled by the width and height of the kernel matrix. 
Thus, for input features $\mathbf{x} = (x_1, x_2, \ldots, x_d)$ and target $y$, the optimization objective of the neural network $f$ in pre-mask mode with parameters $\mathbf{\theta}$ is given by 
\begin{eqnarray}\label{eq:loss_pre_mask_net}
    \underset{\mathbf{\theta}}{\text{argmin}} \;
    \frac{1}{N} \sum_{i=1}^{N} \left(y_i - f(\mathbf{x}_i; \theta) \right)^2 
    + \lambda \cdot \frac{||\, vec(W_1) \, ||_1} {d^2}
\end{eqnarray}
where $N$ denotes the number of samples and $\lambda$ is a regularization parameter. 

\subsection{Masking Vector Extraction and Thresholding}

After completing the pre-masking phase of training, we automatically extract the masking vector for the next training phase. 
The purpose of the binary masking vector is to encode physical relationships between the input features $(X_1, X_2, \ldots, X_d)$ and the target variable $Y$. 
Consequently, we construct a $d$-dimensional vector that identifies variables related to the target among the input features. 
In the first step of creating the masking vector, the vector $\mathbf{m}$ is extracted from the columns in the input layer weight matrix $W_1 = (\mathbf{w}_1 \, \mathbf{w}_2 \ldots \mathbf{w}_d)$ as 
\begin{eqnarray}\label{eq:masking_vector}
\mathbf{m} := [\; || \,\mathbf{w}_1 \, ||_2, || \,\mathbf{w}_2 \, ||_2, \ldots, || \,\mathbf{w}_d \, ||_2 \; ]^T.
\end{eqnarray}
This derivation of the masking vector is similar to previous work \citep{zheng2020, kyono2020a}, but it is more straightforward due to focusing only on the input-to-output relations instead of also considering relations between inputs. 
The resulting vector $\mathbf{m}$ encodes the signal strength between each input and the output. 
As shown in Sec.~\ref{sec:experiments}, it effectively captures physical relationships. 
Next, we binarize $\mathbf{m}$ via thresholding, preventing the network from using input features with low signal strength for prediction. 
Additionally, thresholding helps to reduce the number of false discoveries \citep{zheng2020}. 
The key challenge here is identifying a suitable threshold level.
To address this, we suggest two approaches: 
First, if accessible, (near) ground truth or expert knowledge can guide the choice of threshold. 
In the absence of such information, we propose fine-tuning the network in mask mode across various threshold values, evaluating its predictive accuracy, and selecting the best-performing threshold in terms of training loss. 
We hypothesize that, at this threshold, no essential direct physical drivers are omitted. 
One might expect that the lowest threshold values, which allow most inputs to pass through, would result in the best predictive performance as they allow more information into the network, including possibly spurious, non-physical information. 
However, our experiments consistently show that networks with the highest performance use larger thresholds, even though the differences in training loss are small (not shown).  
This indicates that using only the physical drivers of a process as network inputs is sufficient for accurately predicting the process output. 
Once the threshold value is selected, values in the masking vector below the threshold are assigned a value of zero, while all remaining values are set to one.

\subsection{Physically Consistent Masking and Fine-tuning}

After thresholding the masking vector, we continue training with fine-tuning the model weights in the masking phase, which we refer to as running the model in mask mode. 
The flow of information is illustrated by the red lines in Fig.~\ref{fig:pcm_model}.
In mask mode, instead of passing the inputs through a traditional input layer, we perform element-wise multiplication of the input vector with the thresholded binary masking vector. 
This masks any unrelated input features from influencing the network's output. 
Consequently, the sparsity regularization is now omitted from the optimization objective, leaving it solely defined by the mean squared error of the residuals: 
\begin{eqnarray}\label{eq:loss_mask_net}
    \underset{\mathbf{\theta}}{\text{argmin}} \;
    \frac{1}{N} \sum_{i=1}^{N} \left(y_i - f(\mathbf{x}_i \odot \mathbf{m};  \theta) \right)^2 
\end{eqnarray}
for a model $f$ with parameters $\theta$ and number of samples $N$, where $\odot$ denotes element-wise multiplication.

Overall, three key aspects define the PCMasking framework: 
1) the flexibility to replace the hidden layer block shown in Fig.~\ref{fig:pcm_model} with a different network architecture; 
2) its user-friendly and efficient operation; 
3) its independence from information about physical mechanisms. 
The capability to interchange network architectures makes the PCMasking framework a versatile tool for enhancing existing models with the capability of physical driver selection. 
The automation of both masking vector extraction and thresholding, along with clear guidelines on threshold selection, ensures the PCMasking framework's efficiency and ease of use.  
As for the framework's independence from physical process information, we demonstrate in the following section that network models within the PCMasking framework are nevertheless capable of learning physically consistent connections between inputs and outputs.


\section{Experiments}
\label{sec:experiments}

\begin{figure*}[tb]
    \centering
    \includegraphics[width=\linewidth]{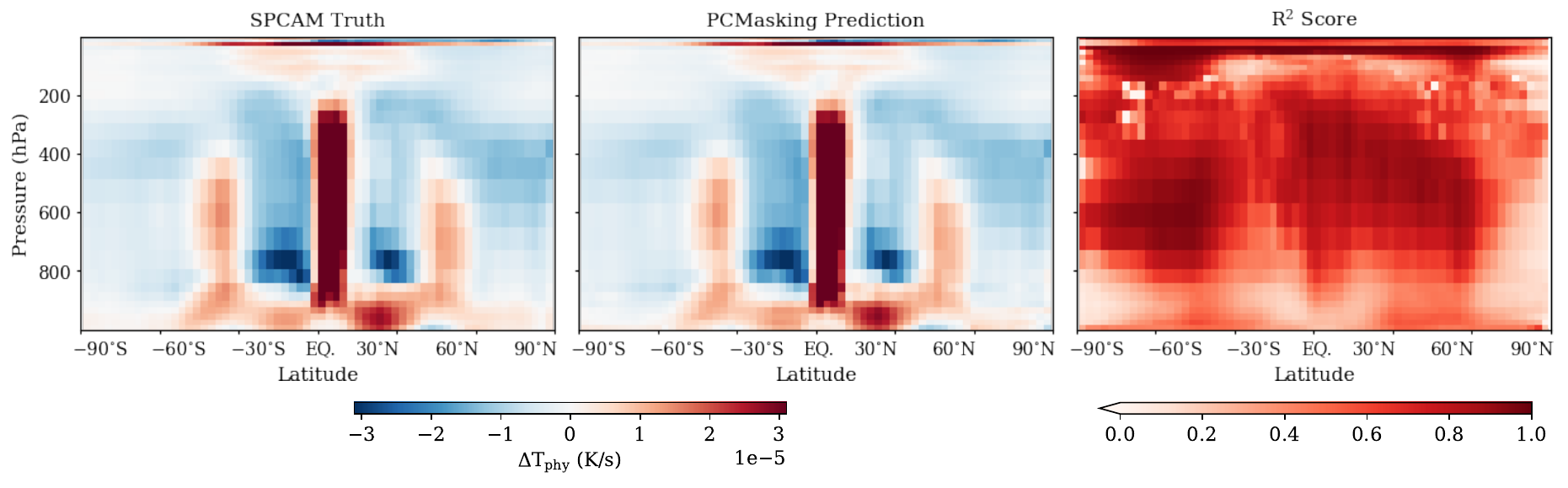}
    \vspace{-6mm}
    \caption{
    Pressure-latitude cross-sections for heating tendencies $\Delta T_{phy}$ computed from 1440 test data samples. 
    One neural network is trained for each of the 30 vertical levels to construct a full vertical profile (y-axis). 
    Each network predicts heating tendencies across the entire globe (x-axis).     
    The left and middle plots illustrate that the PCMasking framework networks accurately predict the true SPCAM values. 
    The right plot depicts the $R^2$ score (higher is better, maximum 1). 
    While the $R^2$ score is high at around 600~hPa in some regions at the equator and the mid-latitudes, the predictive performance declines in the lower troposphere (around 700-1000~hPa). 
    This is likely due to turbulent and stochastic processes in the planetary boundary layer.     
    See SI Fig.~\ref{fig:si_phq_cross_section} for results for moistening tendencies. 
    }
    \label{fig:tphystnd_cross_section}
    \vspace{-2mm}
\end{figure*}

\begin{figure}[tb]
    \centering
    \includegraphics[width=\linewidth]{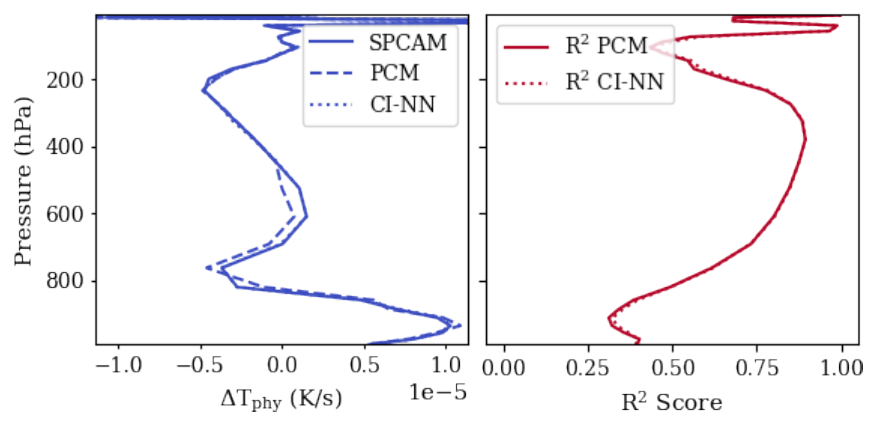}
    \vspace{-5mm}
    \caption{Vertical profiles for heating tendencies $\Delta T_{phy}$ computed from 1440 test data samples. 
    One neural network is trained for each of the 30 vertical levels to construct a full vertical profile (y-axis). 
    The network predictions are horizontally averaged across latitudes. 
    The predictions from the PCMasking framework (PCM) and the causally-informed NNs \citep{iglesias-suarez2024} (CI-NN) are shown on the left alongside the true SPCAM values. 
    Both network types accurately reproduce the true profile. 
    The right plot depicts the $R^2$ score (higher is better, maximum 1).
    The noticeable decline in the lower troposphere (around 700-1000~hPa) is likely due to turbulent and stochastic processes in the planetary boundary layer. 
    See SI Fig.~\ref{fig:si_phq_profile} for results for moistening tendencies.}
    \label{fig:tphystnd_profile}
    \vspace{-2mm}
\end{figure}

\begin{figure*}[tb]
    \centering
    \includegraphics[width=\linewidth]{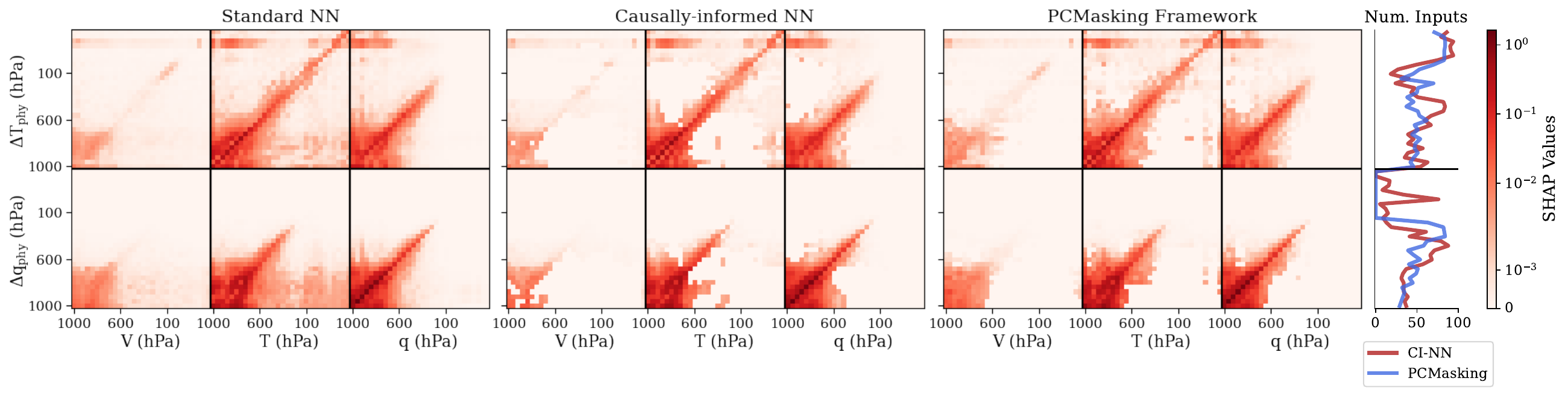}
    \caption{Mean absolute SHAP values computed from 1000 samples for standard feed-forward neural networks (NNs) (left), causally-informed NNs \citep{iglesias-suarez2024} (CI-NN) (middle) and PCMasking framework NNs (right). 
    For clarity, we have only included 3D input and output variables (see SI Fig.~\ref{fig:si_shap_values} for SHAP plots including 2D variables). 
    The standard NNs display numerous spurious connections between input and output variables. 
    This is particularly evident for the input temperature, where inputs in the upper troposphere (around 100-300 hPa) and stratosphere (above 100 hPa) impact outputs in the mid to lower troposphere (around 300-1000 hPa).     
    Such spurious, non-physical links are removed in both the causally-informed NNs and the PCMasking framework. 
    Furthermore, the total number of inputs (right line plot) for CI-NN and the PCMasking framework indicates that the PCMasking framework is more physically accurate as it does not detect any inputs for moistening tendencies in the stratosphere, where the air is cold and dry. 
    }
    \label{fig:shap_values}
    \vspace{-2mm}
\end{figure*}

We empirically evaluate the PCMasking framework's offline performance, i.e., how well the neural networks fit the simulation data, using high-resolution data from the Superparameterized Community Atmosphere Model v3.0 (SPCAM) \citep{collins2006a}. 
This dataset and climate model were also utilized in \citep{iglesias-suarez2024}, providing a baseline comparison for our PCMasking framework. 
\citet{iglesias-suarez2024} employ constraint-based causal discovery \citep{runge2019b} as a pre-step to determine the direct causes of each variable. 
Subsequently, they construct a single-output neural network for each variable, including only the identified causes as network inputs. 
Although this approach may also result in interpretable DL-based parameterizations, the causal discovery method used is computationally expensive, relies on causal assumptions, and introduces its own hyper-parameters that require expert domain knowledge to tune.
In this section, we demonstrate that the PCMasking framework enables us to achieve comparable offline prediction accuracy and physical consistency as \citep{iglesias-suarez2024} while being more efficient and largely automated. 
In fact, even when both methods are already tuned, the PCMasking framework is still about three times more efficient in terms of resource consumption compared to \citet{iglesias-suarez2024} (see Supporting Information (SI) Text~S1 for details). 
In the following section, we will briefly outline the SPCAM setup and the neural network configuration, followed by a presentation of our experimental results.

\subsection{SPCAM and Neural Network Configuration}

SPCAM is a high-resolution global circulation model composed of the Community Atmosphere Model (CAM) with an embedded superparameterization (SP) component -- a 2D storm-resolving model (SRM) within each grid column. 
This embedded SRM, the SP component, features eight north-south oriented 4 km-wide columns and explicitly resolves the majority of deep convective processes while relying on parameterizations for turbulence and microphysics. 
Following \citep{iglesias-suarez2024}, we use SPCAM data from an aquaplanet setup with fixed sea surface temperatures, a realistic equator-to-pole gradient \citep{andersen2012}, and a diurnal cycle without seasonal changes.
The model spans from the surface to the upper stratosphere at $3.5$ hPa, encompassing 30 vertical levels and a horizontal grid resolution of $2.8^{\circ}$ in both latitude and longitude. 
CAM operates with a time step of 30 minutes, while the embedded SRM uses a time step of 20 seconds. 
For further information on SPCAM and the climate model setup, we refer the reader to \citep{iglesias-suarez2024} and its Supporting Information.

\paragraph{SPCAM Data} 
The neural network is tasked to learn the subgrid-scale processes at each time step, as represented by the SP component, based on the atmospheric state provided by the general circulation model (CAM). 
The training dataset covers the SP subgrid resolution of convection and radiation, though with some omissions (e.g., condensates).

\begin{table}[tb]
    \centering
    \caption{Column-wise neural network inputs and outputs. For each 3D variable, $p$ ranges from 0 to 29. }
    \label{tab:inputs_outputs}
    \begin{tabular}{ll}
        \toprule
        \textbf{Variable} & \textbf{Description} \\
        \midrule
        \multicolumn{2}{c}{\textbf{Inputs}} \\
        \midrule
        $T(p)$           & Temperature  (3D) \\
        $q(p)$           & Specific humidity (3D)\\
        $V(p)$           & Meridional wind  (3D)\\
        $P_{srf}$        & Surface pressure \\
        $Q_{sol}$        & Incoming solar radiation at top of atmosphere \\
        $Q_{sen}$        & Sensible-Heat flux at surface \\
        $Q_{lat}$        & Latent-Heat flux at surface \\
        \midrule
        \multicolumn{2}{c}{\textbf{Outputs}} \\
        \midrule
        $\Delta T_{phy}(p)$  & Heating tendencies (3D)\\
        $\Delta q_{phy}(p)$  & Moistening tendencies (3D) \\
        $Q_{sw}^{top}$    & Net shortwave radiative heat flux at top of atmosphere \\
        $Q_{lw}^{top}$   & Net longwave radiative heat flux at top of atmosphere \\
        $Q_{sw}^{srf}$   & Net shortwave radiative heat flux at surface \\
        $Q_{lw}^{srf}$   & Net longwave radiative heat flux at surface \\
        $P$              & Surface precipitation \\
        \bottomrule
    \end{tabular}
    \vspace{-1mm}
\end{table}

Tab. \ref{tab:inputs_outputs} lists the neural network inputs and outputs. 
The input and output variables are arranged into vectors of lengths 94 and 65, respectively. 
It is important to note that in masking mode, only a fraction of the inputs are actually processed by the network. 
The input values are standardized, and the outputs are normalized to ensure a similar order of magnitude (see SI Tab.~\ref{tab:si_variables}). 
We use simulation data from SPCAM spanning three months each for training, validation, and testing, yielding about 45 million data samples for each dataset. 
The training data is shuffled in both time and space (across grid columns).

\paragraph{Neural Network Configuration} 
For most of the neural network and hyper-parameter configurations, we follow \citep{iglesias-suarez2024} to ensure comparability of our results.  
To identify the physical drivers for each output variable, we construct one single-output neural network per variable, resulting in 65 neural networks in total. 
Each model incorporates the same block of hidden layers, comprising 9 fully connected layers with 256 units each. 
These layers employ Leaky Rectified Linear Unit (ReLU) activation with a negative slope of 0.3.
The initial learning rate is set to $0.001$, which is divided by five every three epochs. 
The training batch size is 1024, and the models are optimized using the ADAM optimizer \citep{kingma2017} with default parameters. 
For validation and testing, the batch size is increased to 8192, covering all grid cells of the aquaplanet.
We carry out hyper-parameter tuning for the sparsity loss weighting coefficient, $\lambda$, in pre-mask mode (see Eq.~\ref{eq:loss_pre_mask_net}). 
Exploring a log-spaced search grid $\{1.0, 0.1, \ldots, 1 \times 10^{-5}\}$, we find that $\lambda=0.001$ yields the best outcomes in both prediction accuracy and physical consistency of the identified relevant input features (see SI Fig.~\ref{fig:si_tuning_results}). 
While we use the same $\lambda$ for all models, the masking vector and the threshold selection are customized for each individual output. 
To determine the threshold for a masking vector $\mathbf{m}$, we fine-tune the neural network in mask mode with 20 distinct threshold values. 
The thresholds are evenly distributed within the interval $[1\times 10^{-4}, p_{70})$, where $p_{70}$ represents the $70^{th}$ percentile of the values in $\mathbf{m}$. 
We round these values to four decimal places. 
Once fine-tuning for each threshold value is complete, we proceed to analyze the training loss for each threshold. 
We then select the model with the best performance for each variable.
Training is carried out on a single NVIDIA A100 Tensor Core GPU equipped with 40 GB memory. 
For comparability with \citep{iglesias-suarez2024}, we also train for 18 epochs in total, which we split evenly into 9 epochs training in pre-mask mode and 9 epochs fine-tuning in mask mode. 
The training time in pre-mask mode is roughly 45 minutes for each network (around 0.56 million parameters), while a single fine-tuning run takes approximately 40 minutes per network. 

\subsection{Experimental Results}

\begin{figure}[tb]
    \centering
    \includegraphics[width=\linewidth]{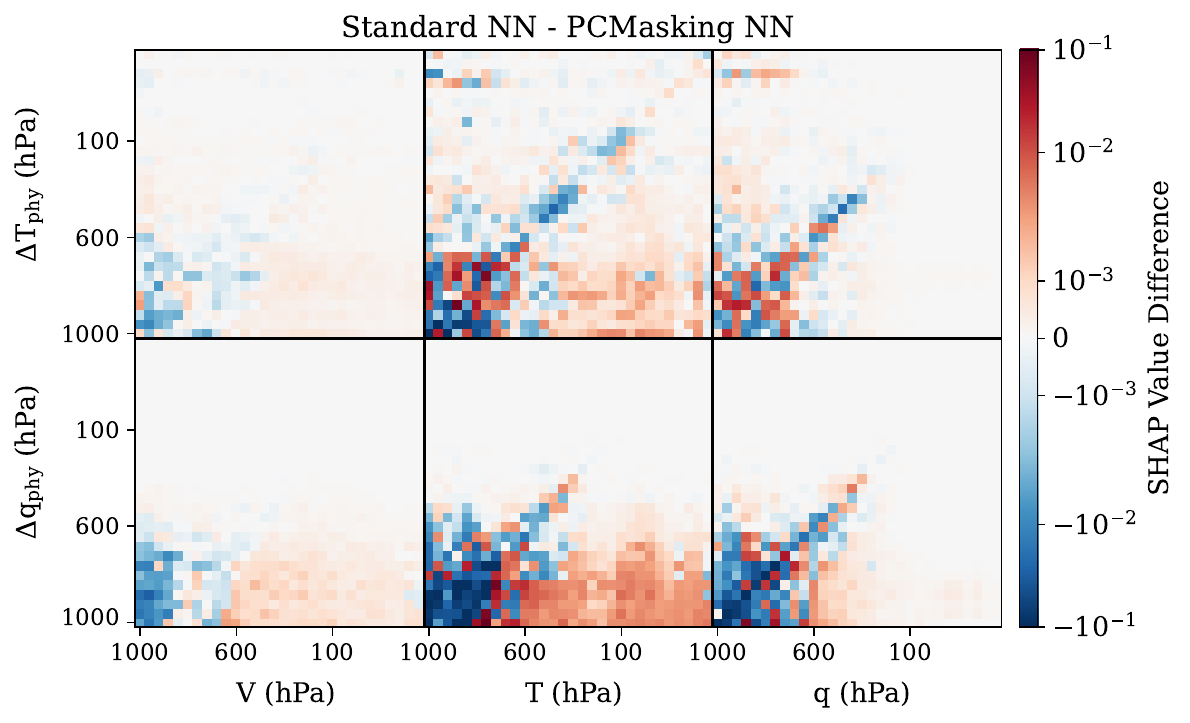}
    \vspace{-4mm}
    \caption{SHAP value difference between mean absolute SHAP values for standard neural networks (NNs) and PCMasking framework networks. 
    SHAP values were computed from 1000 samples, and only 3D variables are displayed for clarity (see SI Fig.~\ref{fig:si_shap_diff} for SHAP difference including 2D variables). 
    {\color{myred}Red} areas indicate positive values, meaning that these input-output links were more pronounced in the standard NNs. 
    Negative, {\color{myblue}blue}  values indicate these connections are more prominent in the PCMasking framework. 
    The PCMasking framework clearly emphasizes physically consistent local interactions along the diagonal and non-local interactions in the lower troposphere (around 700-1000 hPa).}
    \label{fig:shap_diff}
    \vspace{-3mm}
\end{figure}

To determine the PCMasking framework's suitability for physically consistent data-driven climate model parameterization, we evaluate its offline performance on the SPCAM test dataset in terms of both predictive performance and physical consistency.

\paragraph{Offline Predictive Performance} 
Fig.~\ref{fig:tphystnd_cross_section} shows pressure-latitude cross-sections of the average true (left) and predicted (middle) heating tendencies across 1440 samples (about one month). 
The neural networks of the PCMasking framework effectively capture the heating tendencies in terms of horizontal and vertical structure.
They accurately represent key features at the correct geographical locations, e.g., the Intertropical Convergence Zone (ITCZ), evident as the dark red column at the equator, and the heating patterns of mid-latitude storm tracks, depicted in light red at about $\pm 50^\circ$ latitude.  
The horizontal lines at the top of the atmosphere are mostly related to incoming solar radiation. 

In order to examine the predictive performance more closely, we turn to the coefficient of determination, $\mathrm{R}^2$, calculated as one minus the ratio of the residual sum of squares and the total sum of squares. 
The statistical measure serves as a goodness-of-fit quantifier, indicating the extent to which the variance in the dependent variable (the model output $\mathbf{y}$) can be explained by the independent variable (the model input $\mathbf{x}$).  
An $\mathrm{R}^2$ score of 1 means that the predicted values exactly match the observed ones, showing that the model accounts for all variability in the data. 
Conversely, an $\mathrm{R}^2$ score of 0 implies that the model does no better than simply predicting the mean.


The right plot in Fig. \ref{fig:tphystnd_cross_section} displays a pressure-latitude cross-section of the $\mathrm{R}^2$ score for heating tendencies computed from 1440 samples. 
It reveals patches with particularly strong predictive skills near 600~hPa at the equator and in the mid-latitudes, aligning with the locations of the ITCZ and mid-latitude storm tracks. 
However, the predictive performance noticeably declines in the lower troposphere (around 700-1000~hPa), particularly within the planetary boundary layer. 
This reduced performance is likely associated with turbulent and stochastic processes in the planetary boundary layer, leading to increased noise and stochasticity in the data, as previously documented in studies such as \citep{gentine2018a, iglesias-suarez2024, behrens2024}. 
By design, this stochasticity cannot be captured by a deterministic neural network, which provides smoothed-out predictions (see SI Fig.~\ref{fig:si_cross_section_snap_shots}).
The performance drop-off in the lower troposphere is also visible in Fig.~\ref{fig:tphystnd_profile}. 
It shows the horizontal averages of the SPCAM truth, along with the predictions of the causally-informed neural networks (CI-NN) \citep{iglesias-suarez2024} and the PCMasking framework (PCM), as well as the corresponding $\mathrm{R}^2$ scores. 
The performance of CI-NN is equivalent to that of a standard fully connected feed-forward neural network (not shown, see \citep{iglesias-suarez2024}). 
Our predictive performance closely aligns with that of CI-NN, including similar limitations in the lower atmosphere where reliable prediction is particularly challenging due to stochasticity, as previously discussed. 
However, the PCMasking framework consumes about two-thirds fewer resources than CI-NN, making it substantially more efficient while maintaining nearly the same level of performance.


\paragraph{Physical Consistency and Interpretability} 
For evaluation in terms of interpretability and physical consistency, we turn to SHapley Additive exPlanations (SHAP) \citep{lundberg2017}, a framework for explaining the output of machine learning models. 
Based on the concept of Shapley values from game theory, SHAP values measure the contribution of each feature to the prediction for each data sample, thus providing insight into how each feature influences the model's decision-making process.
Fig.~\ref{fig:shap_values} depicts the mean absolute SHAP values for a vanilla fully connected feed-forward neural network (left), CI-NN \citep{iglesias-suarez2024} (middle), and our PCMasking framework (right). 
The plots show which input variables (on the x-axis) the networks are using for the prediction of the output variables (on the y-axis). 
The values are computed from 1000 data samples. 
For clarity, we are only presenting 3D input and output variables and refer to the SI for SHAP plots that include 2D variables.

Based on physical knowledge about the climate system, we know that in the lower troposphere, interactions are generally both local and non-local due to mixing in the planetary boundary layer and buoyancy plumes from the surface. 
Conversely, in the upper troposphere (around 100-300~hPa) and stratosphere (above 100~hPa), processes are predominantly local.
The left plot in Fig.~\ref{fig:shap_values} clearly illustrates that the vanilla feed-forward NNs, which utilize all inputs for predictions, exhibit numerous connections between inputs throughout the atmosphere, particularly between heating and moistening tendencies in the lower troposphere and temperature in the stratosphere. 
Such non-local interactions conflict with our physical understanding, i.e., convection is mainly driven by processes within the troposphere, such as adiabatic cooling and heating, cloud formation, and latent heat release during phase changes of water. 
This suggests that these are spurious connections likely learned due to vertical correlation in the atmosphere due to convective processes.
In contrast, the middle and right plots in Fig.~\ref{fig:shap_values} both demonstrate that CI-NN and our PCMasking framework successfully remove these spurious links. 
This is further highlighted in the SHAP difference between the vanilla NN and the PCMasking framework in Fig.~\ref{fig:shap_diff}. 
The elimination of spurious links in the PCMasking framework is particularly evident for input variable temperature, as indicated by the positive, red areas in Fig.~\ref{fig:shap_diff}. 
Furthermore, this removal of spurious connections in the PCMasking framework results in a greater focus on the connections between subgrid-scale processes and the actual physical drivers. 
This is evidenced by the negative, blue areas for both non-local and local interactions in the lower troposphere as well as for local interactions along the diagonal.

Moreover, the right side of Fig.~\ref{fig:shap_values} shows the number of inputs identified by CI-NN and the PCMasking framework at each vertical level for moistening and heating tendencies. 
Although the overall patterns of peaks and valleys are similar, there is a notable difference: 
The PCMasking framework successfully avoids identifying any input variables in the stratosphere for moistening tendencies. 
This is expected due to the dry air and the absence of most convective processes in the stratosphere.  
Overall, the SHAP value comparison between the three types of networks demonstrates that 
1) both CI-NN and the PCMasking framework effectively remove non-physical, spurious links, and 2) in doing so, networks within the PCMasking framework focus on actual physical connections in their predictions. 


\paragraph{Evaluation on Different Climates}

\begin{figure}[tb]
    \centering
    \includegraphics[width=\linewidth]{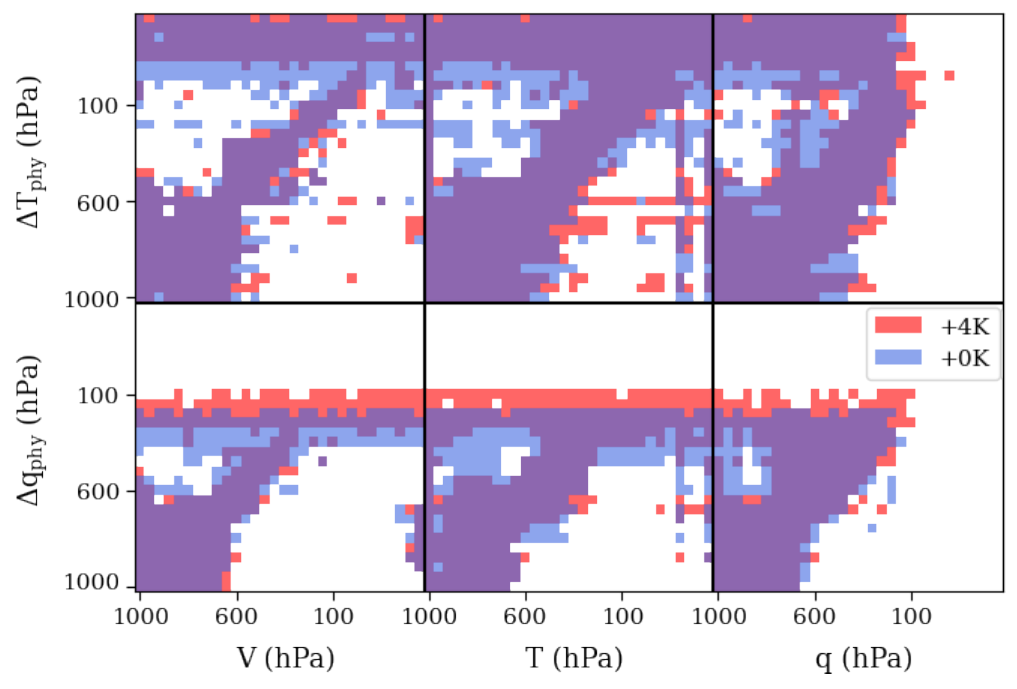}
    \vspace{-4mm}
    \caption{Physical drivers for moistening and heating tendencies found with the PCMasking framework. 
    Selected inputs for the 0~K reference climate are shown in blue and for the +4~K climate in red.     
    {\color{myviolet}Violet} areas indicate where the physical drivers overlap. 
    While physical drivers in the lower troposphere (around 700-1000 hPa) remain largely unchanged, there is a noticeable upward shift for moistening tendencies at around 100 hPa for +4~K. 
    This is consistent with convective processes occurring at higher altitudes in warmer climates. 
    See SI Fig.~\ref{fig:si_inputs_plus4k} and \ref{fig:si_inputs_minus4k} for a comparison of 0~K and -4~K climates as well as 2D variables. 
    }
    \label{fig:climate_plus4k}
    \vspace{-3mm}
\end{figure}
In order to demonstrate the ability of our PCMasking framework to consistently identify physical drivers across different climates, we trained on different SPCAM simulations with sea surface temperatures of +4 Kelvin (K) and -4~K compared to the reference climate (+0~K). 
Fig.~\ref{fig:climate_plus4k} presents the number of inputs found in both the reference climate (+0~K) and the warmed climate (+4~K).
While local and non-local drivers in the lower troposphere remain largely the same across both climates, there is a noticeable upward shift of moistening tendencies, occurring at approximately 100~hPa. 
This is consistent with deep convective processes occurring at higher altitudes in warmer climates, further providing strong evidence that our PCMasking framework is indeed identifying real-world physical drivers.
However, the slight change in physical drivers also indicates that achieving good generalization across different climates using this methodology is limited and warrants more thorough exploration of generalization performance in future work.


\section{Conclusion}
\label{sec:conclusion}
 
In this work, we introduce the PCMasking framework, an efficient training scheme that results in physically consistent and interpretable neural network models specifically designed for climate model parameterization. 
We have demonstrated empirically that the PCMasking framework reliably identifies real-world physical drivers of convective processes across different climate conditions without explicitly relying on physical constraints. 
Furthermore, networks trained within the PCMasking framework are competitive with previous work \cite{iglesias-suarez2024} in terms of offline predictive performance at only a third of the computational cost. 
The PCMasking framework features an automated training process, making it efficient and easy to use. 
Moreover, the interchangeability of its internal architecture renders the PCMasking framework a versatile tool for enhancing predictive models with physical driver selection. 
Possible other applications are forecasting air pollutants or predicting sea ice concentrations in a climate model.  
More broadly, the PCMasking framework may be useful outside of climate science, for example, in identifying and quantifying gene regulatory effects in single-cell RNA measurements.

While this study presents a step towards physically consistent deep neural network models for climate model parameterizations, there are still several challenges that remain to be addressed. 
These include generalization to different climates and online stability, i.e., the stability of hybrid model climate simulations \citep{lin2023}. 
Tackling the problem of generalization, \citet{beucler2024} propose a strategy of transforming the inputs and outputs to maintain similar distributions under different climate conditions. 
This idea could readily be integrated into our PCMasking framework.
Moreover, the transition from offline to online performance is a major challenge in data-driven DL-based parameterizations, as success in offline settings does not always translate to stable coupled model runs \cite{brenowitz2020b, lin2023}. 
\citet{brenowitz2020a} suggest that the numerical instabilities commonly observed in coupled model runs could be related to non-causal correlations or sub-optimal selections of network architecture and hyper-parameters. 
However, non-causal correlations have not been found to be closely related to hybrid model instabilities \citep{iglesias-suarez2024}. 
This is a key challenge in hybrid modeling and remains an active area of research. 

Overall, we demonstrate that the PCMasking framework represents an innovative step in addressing a major challenge of data-driven models by respecting the underlying physical processes of the data. 
Our framework advances the process-based representation of complex phenomena. 
Given its flexibility and applicability to other predictive tasks, it can benefit not just climate science but other scientific disciplines as well. 


\section*{Acknowledgments}

This study was funded by the ERC Synergy Grant USMILE (Grant Agreement 855187) under the Horizon 2020 research and innovation program and by the Horizon Europe project AI4PEX (Grant Agreement 101137682). 
VE was also supported by the Deutsche Forschungsgemeinschaft (DFG) Gottfried Wilhelm Leibniz Prize (Reference No. EY 22/2-1). 
This work used resources from both the Deutsches Klimarechenzentrum (DKRZ) granted by its Scientific Steering Committee (WLA) under project ID 1179 (USMILE) and the JUWELS supercomputer at the Jülich Supercomputing Centre (JSC) under the "Machine learning-based parameterizations and analysis for the ICON model" project.


\printbibliography


\newpage
\appendix

\renewcommand{\thefigure}{S\arabic{figure}}
\renewcommand{\thetable}{S\arabic{table}}

\setcounter{figure}{0}
\setcounter{table}{0}


\title{Supporting Information: Towards Physically Consistent Deep Learning For Climate Model Parameterizations
}

\maketitle
\section*{Code}

Code is available at \url{https://github.com/EyringMLClimateGroup/kuehbacher24ICMLA_PCMasking}. 

\section*{Text S1: Computing Time Comparison}
We compared the computing resources of our PCMasking framework and the causally-informed neural networks (NNs) presented by \citet{iglesias-suarez2024}. 
The run time and resource consumption for all network types (standard feed-forward NN, causally-informed NN \citep{iglesias-suarez2024}, and our PCMasking framework) are about the same. 
Using 4 GPUs on an HPC compute node (GPUs: 4x Nvidia A100 80GB/GPU; CPUs: 2x AMD 7763 CPU, 128 cores in total, 512 GB main memory), training all 65 networks consumes about 25 node-hours (about $0.38$ node-hours per output variable).  

\citet{iglesias-suarez2024} require additional resources for their causal discovery pre-step, using the constraint-based PC$_1$ algorithm of the PCMCI framework \citep{runge2019b}. 
PC$_1$ takes about 46 seconds per grid cell for the SPCAM dataset (2x AMD 7763 CPU; 128 cores in total, 256 GB main memory).
The SPCAM grid has 8192 grid cells (128 longitudes $\times$ 64 latitudes). 
Assuming one can run PC$_1$ perfectly and efficiently using a compute node (128 jobs in parallel with 1 job per core), we estimate the computation for PC$_1$ as: 
65 outputs $\times$ 64 jobs (8192 column / 128 cores) $\times$ 46 seconds $ \approx 53$ additional node-hours (about $0.81$ node-hours per output variable). 

Thus, CI-NN requires 3x more computing resources compared to our PCMasking framework. 

\begin{table*}[!htb]
\centering
\caption{Summary of neural network input and output variables. The values are equivalent to \citep{iglesias-suarez2024}. 
The input variables are standardized, and the output values are normalized.}
\label{tab:si_variables}
\begin{tabular*}{0.95\textwidth}{@{\extracolsep\fill}lll|lll@{\extracolsep\fill}}
\toprule
 & Inputs & Units & Outputs & Units & Normalization \\ 
\midrule
 & Temperature, \( T(p) \) & K & Heating tendencies, \( \Delta T_{phy}(p) \) & \( K s^{-1} \) & \( C_p \) \\
 & Specific humidity, \( q(p) \) & \( kg kg^{-1} \) & Moistening tendencies \( \Delta q_{phy}(p) \) & \( kg kg^{-1} s^{-1} \) & \( L_v \) \\
 & Meridional wind, \( V(p) \) & \( m s^{-1} \) & Net shortwave radiative heat flux at TOA, \( Q_{sw}^{top} \) & \( W m^{-2} \) & \( 10^{-3} \) \\
 & Surface pressure, \( P_{srf} \) & Pa & Net longwave radiative heat flux at TOA, \( Q_{lw}^{top} \) & \( W m^{-2} \) & \( 10^{-3} \) \\
 & Incoming solar radiation, \( Q_{sol} \) & \( W m^{-2} \) & Net shortwave radiative heat flux at the surface, \( Q_{sw}^{srf} \) & \( W m^{-2} \) & \( 10^{-3} \) \\
 & Sensible heat flux, \( Q_{sen} \) & \( W m^{-2} \) & Net longwave radiative heat flux at the surface, \( Q_{lw}^{srf} \) & \( W m^{-2} \) & \( 10^{-3} \) \\
 & Latent heat flux, \( Q_{lat} \) & \( W m^{-2} \) & Precipitation, P & \( kg m^{-2} d^{-1} \) & \( 1.728 \times 10^{6} \) \\
\bottomrule
\end{tabular*}
\end{table*}

\begin{table*}[!htb]
\centering
\caption{Neural network configuration and hyper-parameter settings. The random seed used for training was 42.}
\label{tab:si_network_configuration}
\begin{tabular}{l l}
\toprule
Hidden layers & 9 \\
Units per hidden layer & 256 \\
Hidden layer activation function & Leaky ReLU ($\text{neg. slope } = 0.3$)\\
$\lambda$ & 0.001 \\
Optimizer & ADAM \cite{kingma2017} with default parameters\\
Initial learning rate & 0.001\\
Learning rate schedule & Divide by 5 every 3 epochs\\
Training batch size & 1024 \\
Validation batch size & 8192 \\
Total number of epochs & 18\\
$\quad$ Pre-mask mode & 9\\
$\quad$ Mask mode & 9\\
\midrule
Number of parameters & \\
$\quad$ Pre-mask mode & about 0.56 million\\
$\quad$ Mask mode & about 0.55 million\\
\bottomrule
\end{tabular}

\end{table*}

\begin{figure*}[tb]
    \centering
    \begin{minipage}[t]{0.468\linewidth}
        \centering
        \subcaption{}
        \vspace{-4mm}
        \includegraphics[width=\linewidth, trim={0 2.7cm 2.7cm 0}, clip]{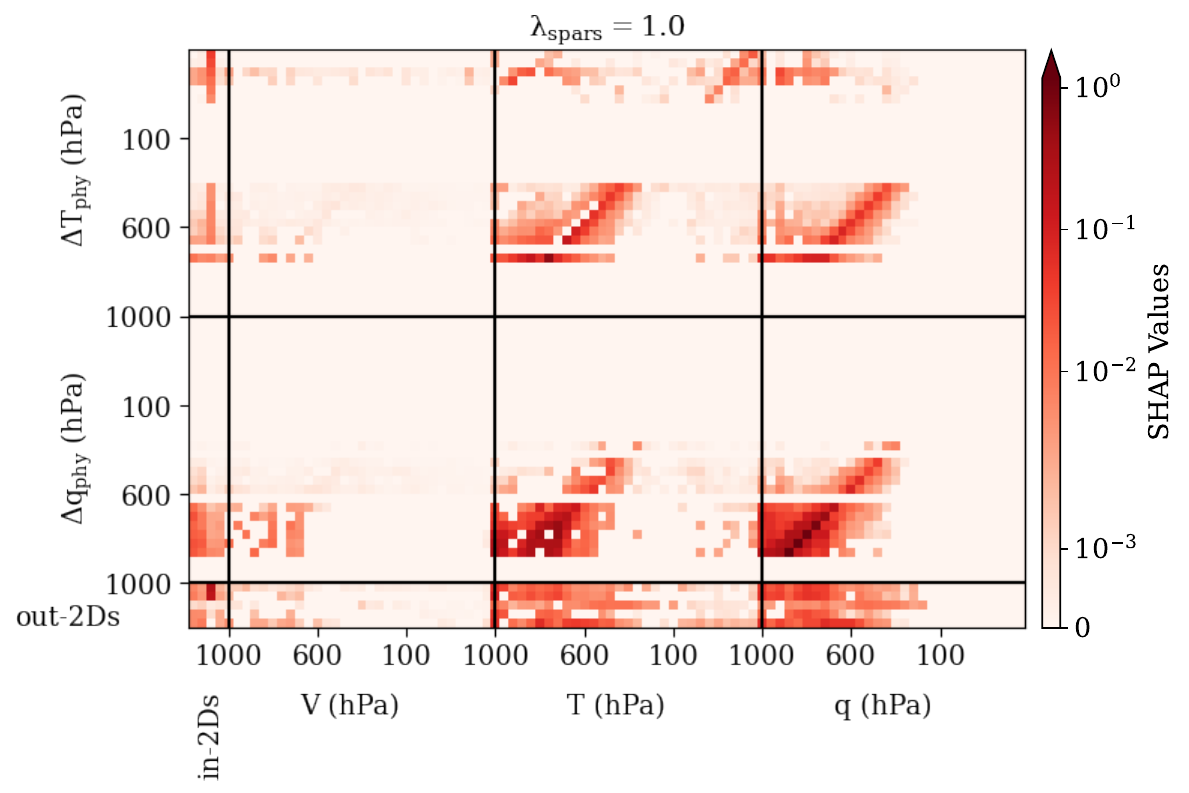}
        \label{fig:shap_spars1.0}
    \end{minipage}
    \hspace{4mm}
    \begin{minipage}[t]{0.46\linewidth}
        \centering
        \subcaption{}
        \vspace{-4mm}
        \includegraphics[width=\linewidth, trim={3cm 2.7cm 0 0}, clip]{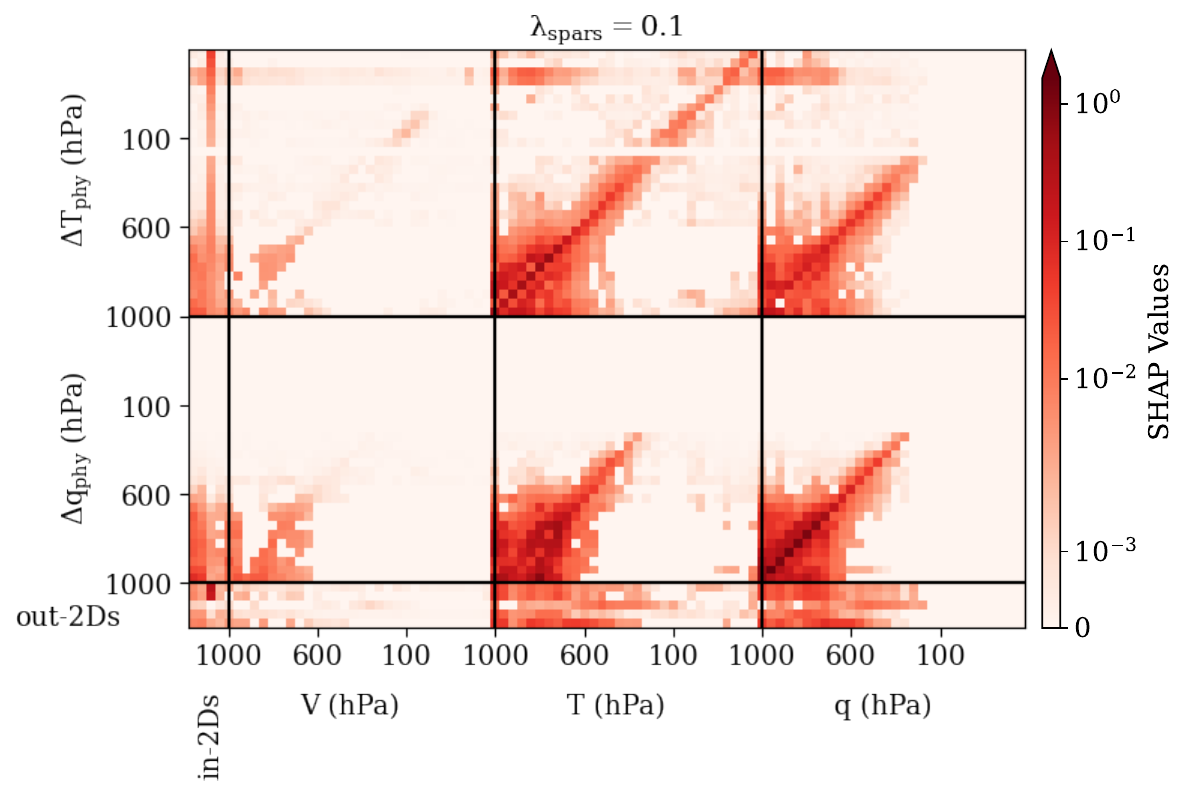}      
        \label{fig:shap_spars0.1}
    \end{minipage}
    \hfill
    \vspace{2mm} 
    \begin{minipage}[t]{0.468\linewidth}
        \centering
        \subcaption{}
        \vspace{-4mm}
        \includegraphics[width=\linewidth, trim={0 2.7cm 2.7cm 0}, clip]{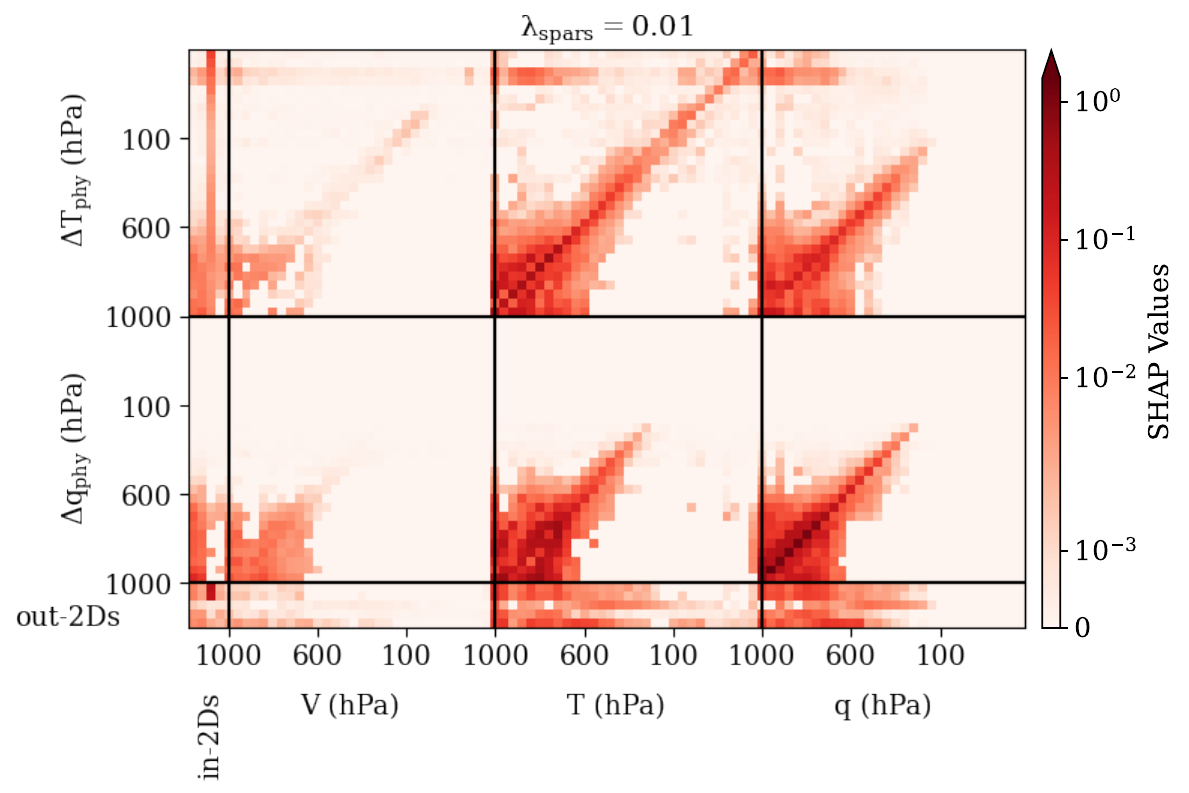}
        \label{fig:shap_spars0.01}
    \end{minipage}
    \hspace{4mm}
    \begin{minipage}[t]{0.46\linewidth}
        \centering
        \subcaption{}
        \vspace{-4mm}
        \includegraphics[width=\linewidth, trim={3cm 2.7cm 0 0}, clip]{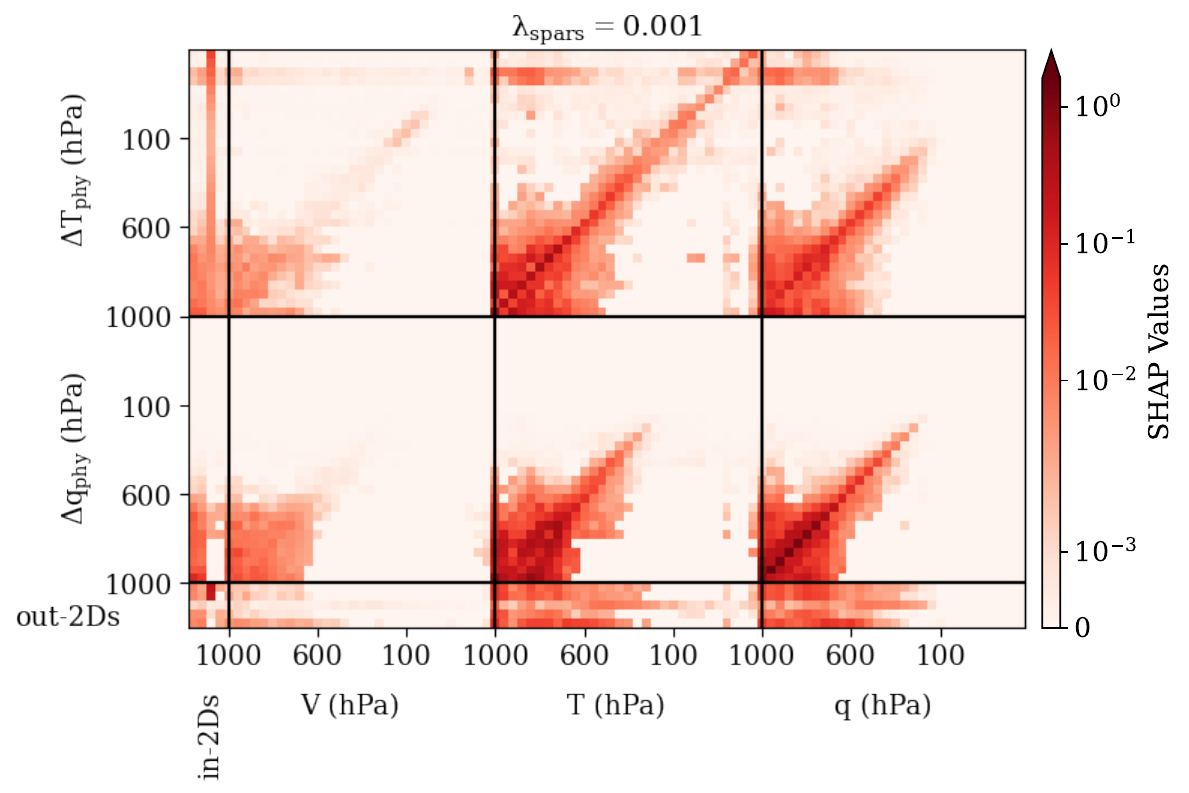}
        \label{fig:shap_spars0.001}
    \end{minipage}
    \hfill
    \vspace{2mm} 
    \begin{minipage}[t]{0.468\linewidth}
        \centering
        \subcaption{}
        \vspace{-4mm}
        \includegraphics[width=\linewidth, trim={0 0 2.7cm 0}, clip]{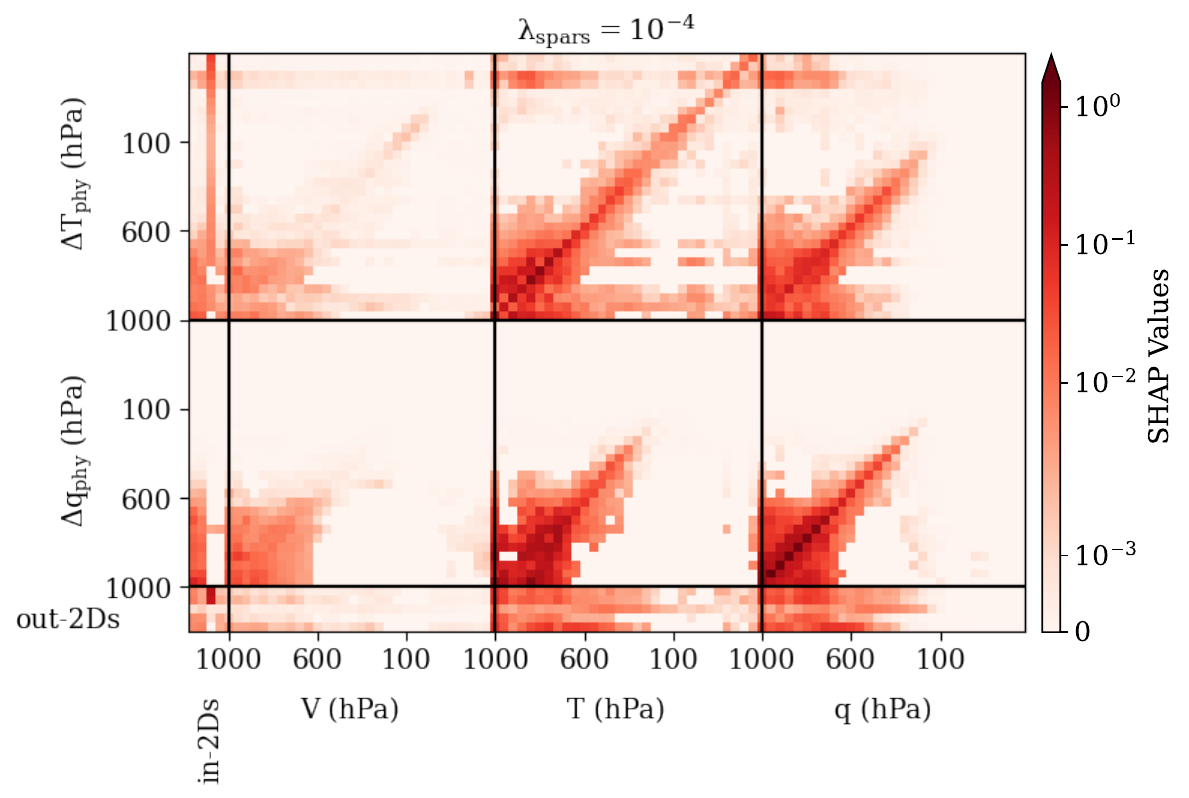}
        \label{fig:shap_spars1e-4}
    \end{minipage}
    \hspace{4mm}
    \begin{minipage}[t]{0.46\linewidth}
        \centering
        \subcaption{}
        \vspace{-4mm}
        \includegraphics[width=\linewidth, trim={3cm 0 0 0}, clip]{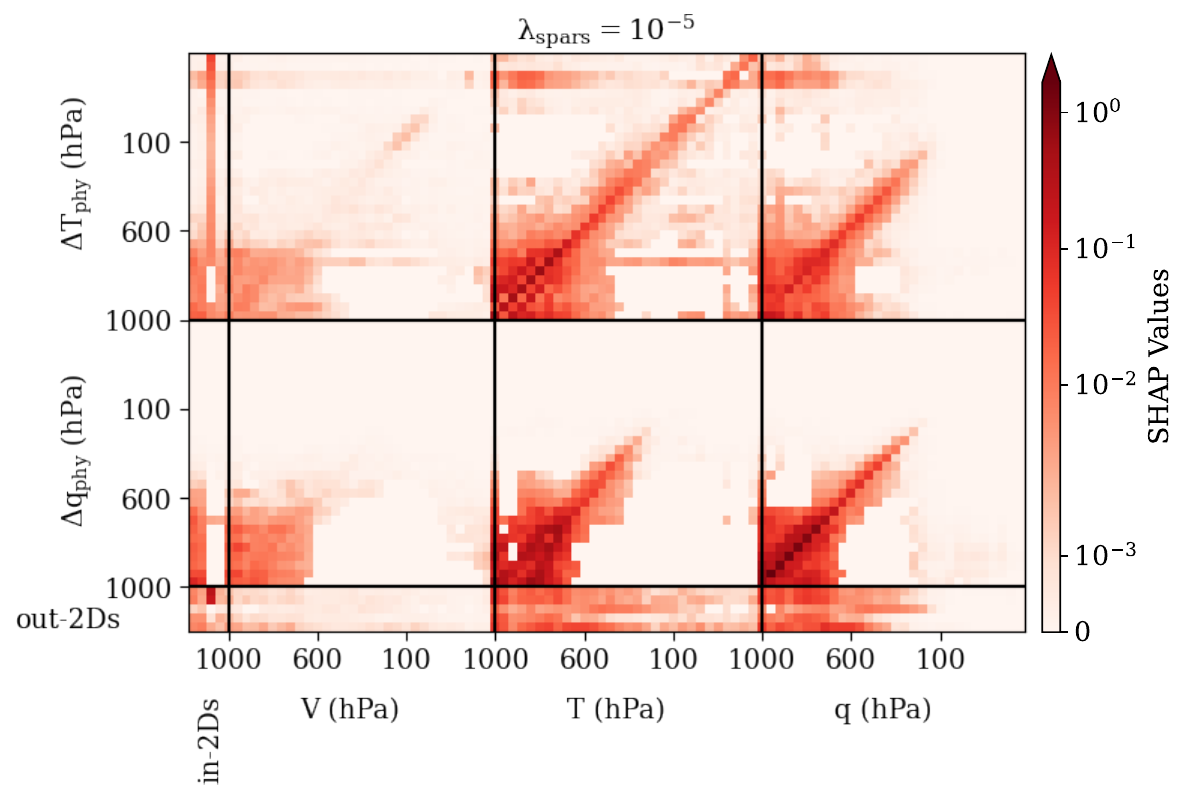}
        \label{fig:shap_spars1e-5}
    \end{minipage}
    \hfill
    \label{fig:tuning_results1}
\end{figure*}

\begin{figure*}[tb]
\renewcommand{\thefigure}{S\arabic{figure}}

    \ContinuedFloat
    \centering
    \begin{minipage}[t]{0.8\linewidth}
        \centering
        \subcaption{}
        \vspace{-4mm}
        \includegraphics[width=\linewidth]{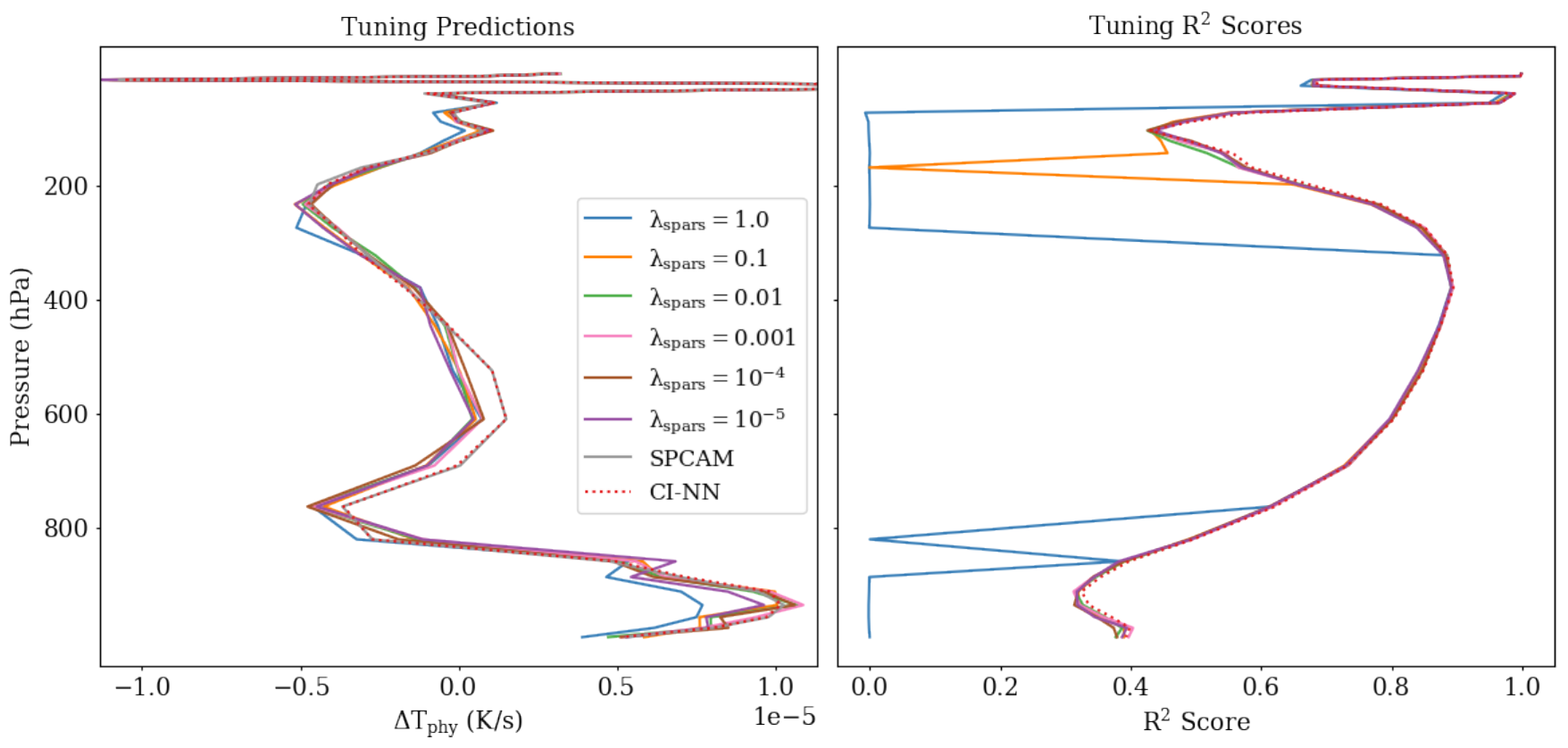}
        \label{fig:tphystnd_tuning_profiles}
    \end{minipage}
    \vspace{2mm}
    \begin{minipage}[t]{0.8\linewidth}
        \centering
        \subcaption{}
        \vspace{-4mm}
        \includegraphics[width=\linewidth]{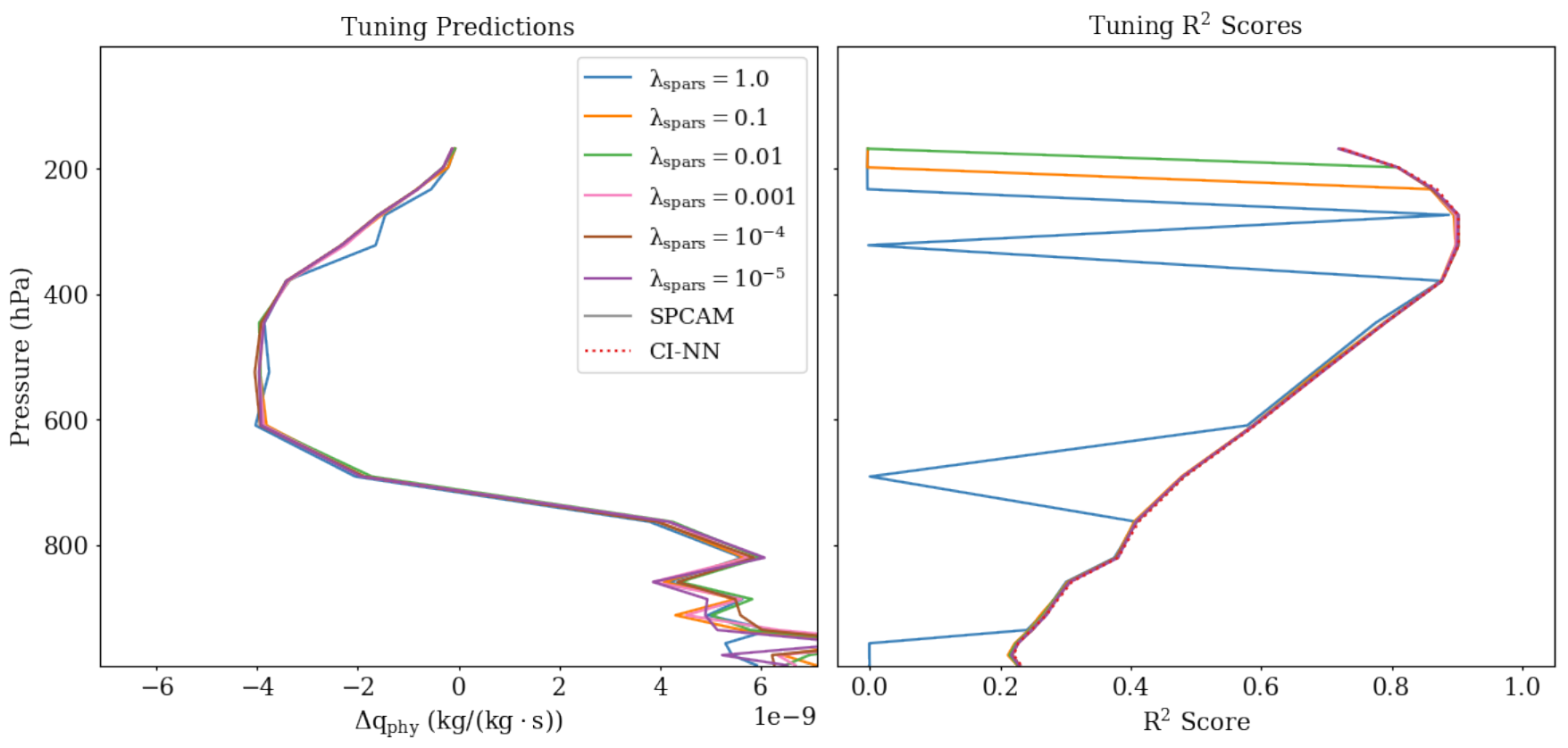}
        \label{fig:phq_tuning_profiles}
    \end{minipage}

    \caption{Tuning results for regularization parameter $\lambda$ in the PCMasking framework exploring a log-spaced search grid $\{1.0, 0.1, \ldots, 1 \times 10^{-5}\}$. 
    To evaluate physical consistency, we consider mean absolute SHAP values computed from 1000 data samples for PCMasking networks trained with 
    $\lambda = 1.0$ (\protect\subref{fig:shap_spars1.0}),  
    $\lambda = 0.1$ (\protect\subref{fig:shap_spars0.1}), 
    $\lambda = 0.01$ (\protect\subref{fig:shap_spars0.01}), 
    $\lambda = 0.001$ (\protect\subref{fig:shap_spars0.001}), 
    $\lambda = 10^{-4}$ (\protect\subref{fig:shap_spars1e-4}), and
    $\lambda = 10^{-5}$ (\protect\subref{fig:shap_spars1e-5}). 
    Prediction accuracy is measured by comparing predictive performance against SPCAM truth and the causally-informed neural network (CI-NN) from \cite{iglesias-suarez2024}, as well as computing the $R^2$ score using 1440 samples. 
    These results are shown for temperature and moistening tendencies in Figures (\protect\subref{fig:tphystnd_tuning_profiles}) and 
    (\protect\subref{fig:phq_tuning_profiles}), respectively.
    We find that the neural networks trained with $\lambda = 0.001$ give the best performance both in terms of physical consistency and performance.
    }
    \label{fig:si_tuning_results}
    \vspace{5mm}
\end{figure*}

\begin{figure*}[tb]
    \centering
    \includegraphics[width=\linewidth]{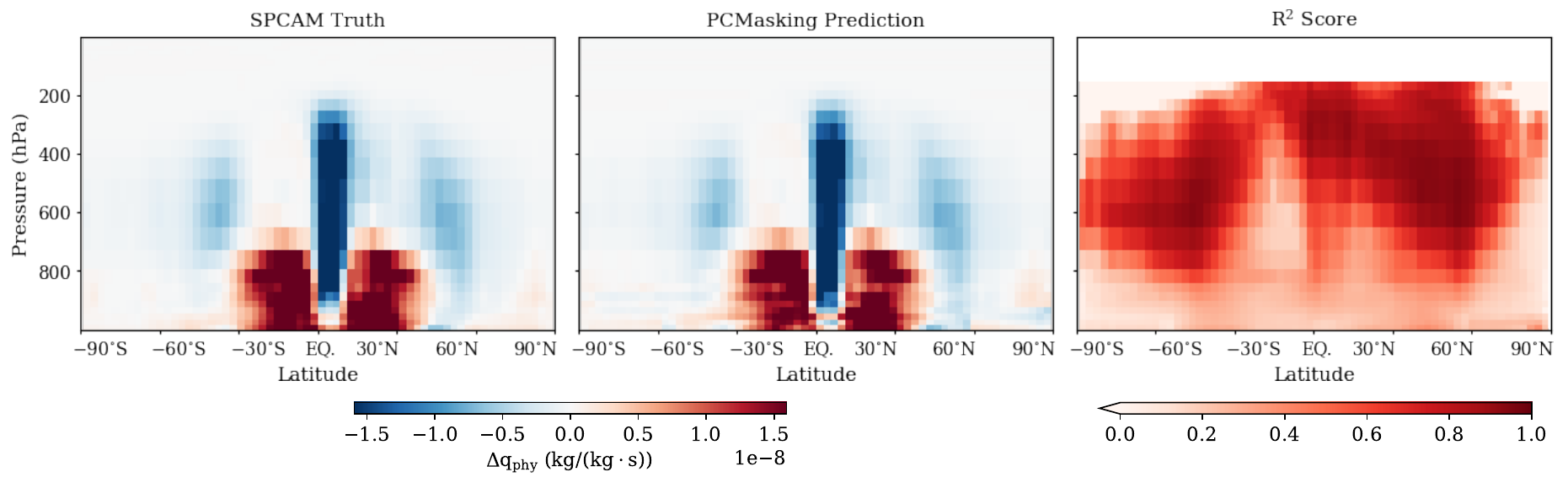}
    \caption{
    Same as Fig. 2 but for moistening tendencies $\Delta q_{phy}$. 
    Pressure-latitude cross-sections computed from 1440 test data samples. 
    One neural network is trained for each of the 30 vertical levels to construct a full vertical profile (y-axis). 
    Each network predicts moistening tendencies across the entire globe (x-axis).    
    The left and middle plots show the true and predicted moistening tendencies, respectively. 
    The right plot depicts the $R^2$ score (higher is better, maximum 1). 
    Negative $R^2$ scores are cut off.  
    }
    \label{fig:si_phq_cross_section}
\end{figure*}

\begin{figure*}[tb]
    \centering
    \includegraphics[width=0.7\linewidth]{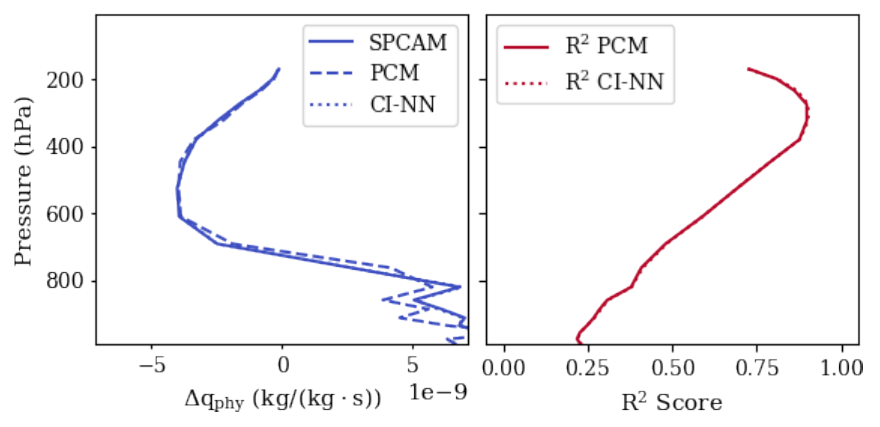}
    \caption{ 
    Same as Fig. 3 but for moistening tendencies $\Delta q_{phy}$. 
    Vertical profiles computed from 1440 test data samples. 
    One neural network is trained for each of the 30 vertical levels to construct a full vertical profile (y-axis). 
    The network predictions are horizontally averaged across latitudes. 
    The predictions from the PCMasking framework (PCM) and the causally-informed NNs \citep{iglesias-suarez2024} (CI-NN) are shown on the left alongside the true SPCAM values. 
    The right plot depicts the $R^2$ scores for both PCM and CI-NN (higher is better, maximum 1). 
    Negative $R^2$ scores are cut off.
   }
    \label{fig:si_phq_profile}
\end{figure*}

\begin{figure*}[ht]

    \centering
    \begin{minipage}[t]{0.45\linewidth}
        \centering
        \subcaption{}
        \vspace{-4mm}
        \includegraphics[width=\linewidth]{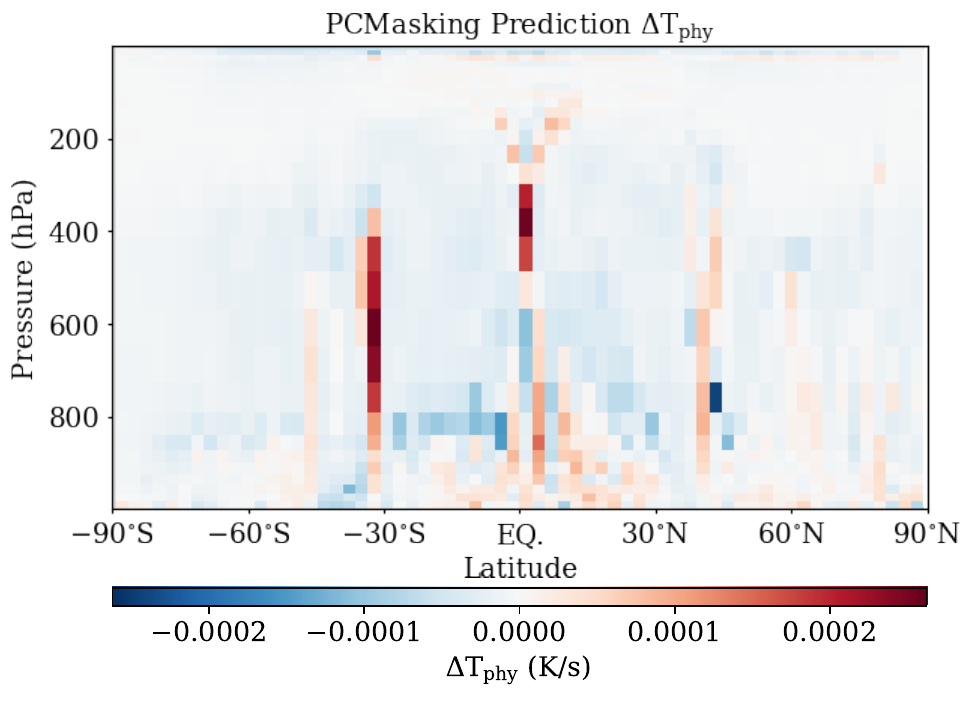}
        \label{fig:tphsytnd_cross_section_snap_shot_pred}
    \end{minipage}
    \hfill
    \begin{minipage}[t]{0.45\linewidth}
        \centering
        \subcaption{}
        \vspace{-4mm}
        \includegraphics[width=\linewidth]{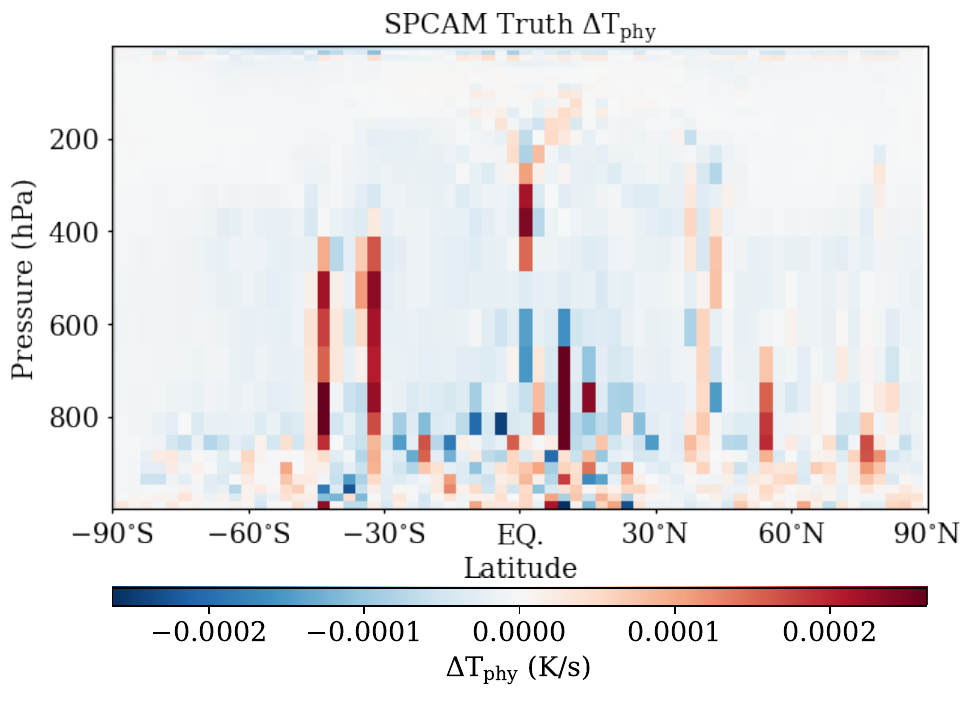}      
        \label{fig:tphsytnd_cross_section_snap_shot_truth}
    \end{minipage}
     \begin{minipage}[t]{0.45\linewidth}
        \centering
        \subcaption{}
        \vspace{-4mm}
        \includegraphics[width=\linewidth]{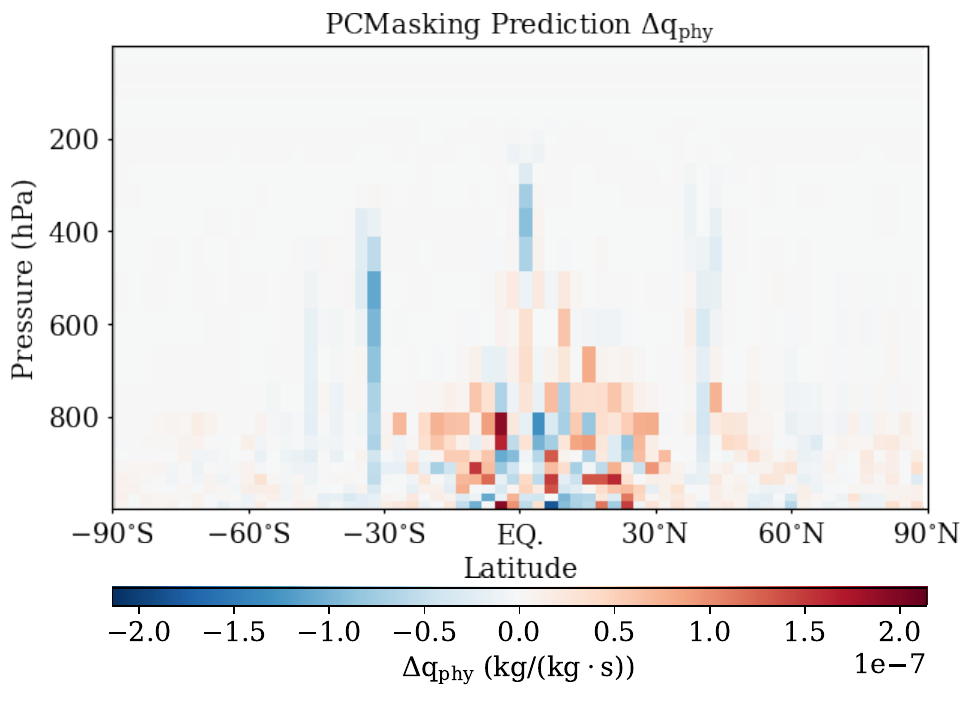}
        \label{fig:phq_cross_section_snap_shot_pred}
    \end{minipage}
    \hfill
    \begin{minipage}[t]{0.45\linewidth}
        \centering
        \subcaption{}
        \vspace{-4mm}
        \includegraphics[width=\linewidth]{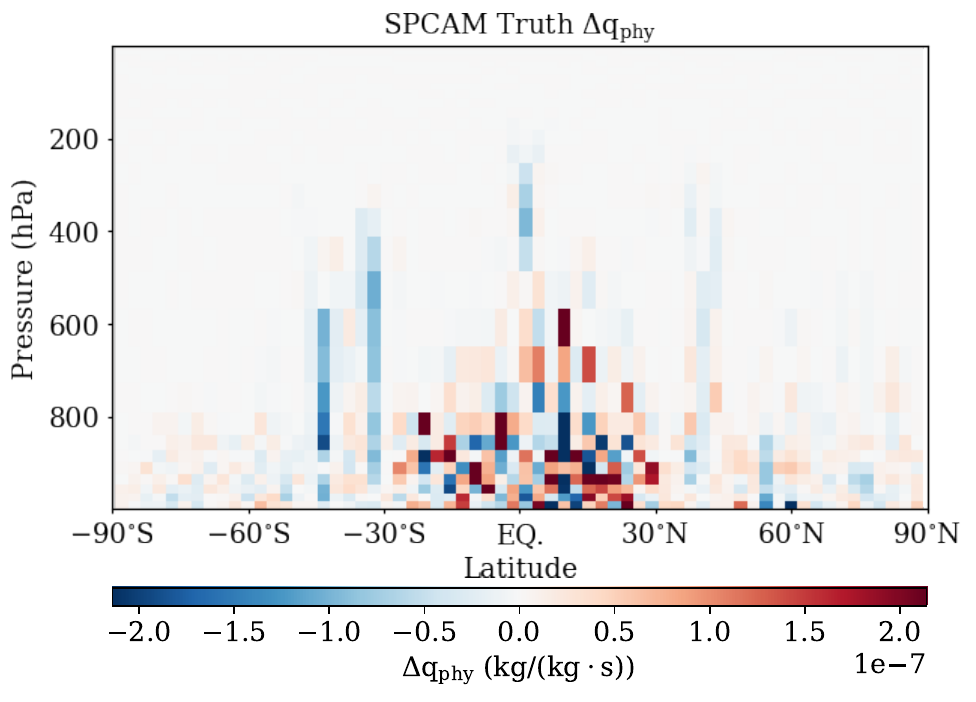}
        \label{fig:phq_cross_section_snap_shot_truth}
    \end{minipage}
    \caption{Pressure-latitude cross-section snapshots for heating and moistening tendencies at time step 500 for PCMasking predictions and SPCAM truth. 
    One neural network is trained for each of the 30 vertical levels to construct a full vertical profile (y-axis).
    Each network predicts values across the entire globe (x-axis). 
    In contrast to Fig.~2 and Fig.~S2, which show averaged predictions, these are single-time step predictions to illustrate smoothing in the NN predictions.  
    While the predicted and true values show the same general pattern, the neural network predictions are considerably smoother and exhibit less extreme values compared to the SPCAM truth. 
    Subfigures (\protect\subref{fig:tphsytnd_cross_section_snap_shot_pred}) and (\protect\subref{fig:tphsytnd_cross_section_snap_shot_truth}) show predicted and true values for heating tendencies $\Delta T_{phys}$, respectively.
    SPCAM truth and network predictions for moistening tendencies $\Delta q_{phys}$ are depicted in subfigures 
    (\protect\subref{fig:phq_cross_section_snap_shot_pred}) and 
    (\protect\subref{fig:phq_cross_section_snap_shot_truth}), respectively. 
    }
    \label{fig:si_cross_section_snap_shots}
\end{figure*}

\begin{figure*}[ht]
    \centering
    \begin{minipage}[b]{0.55\textwidth}
        \centering
        \subcaption{}
        \vspace{-4mm}
        \includegraphics[width=\textwidth]{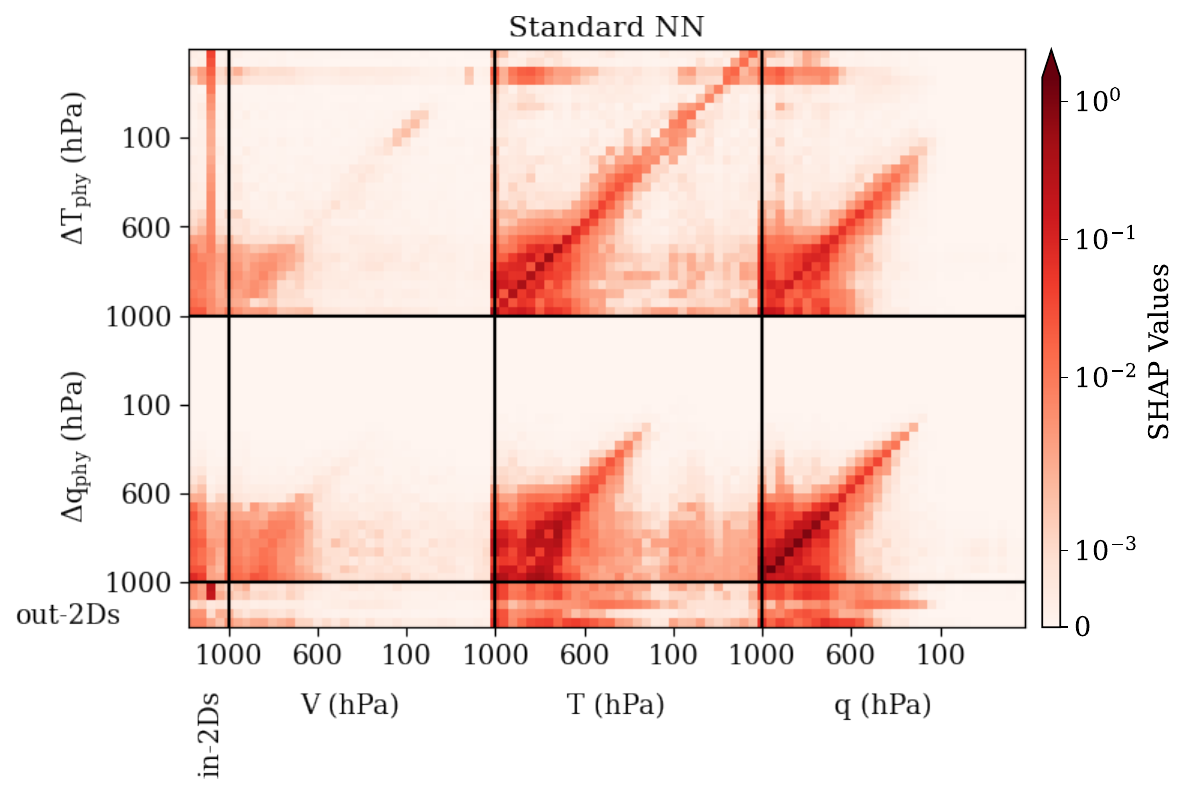}
        \label{fig:si_shap_single}
    \end{minipage}
    
    \begin{minipage}[b]{0.55\textwidth}
        \centering
        \subcaption{}
        \vspace{-4mm}
        \includegraphics[width=\textwidth]
        {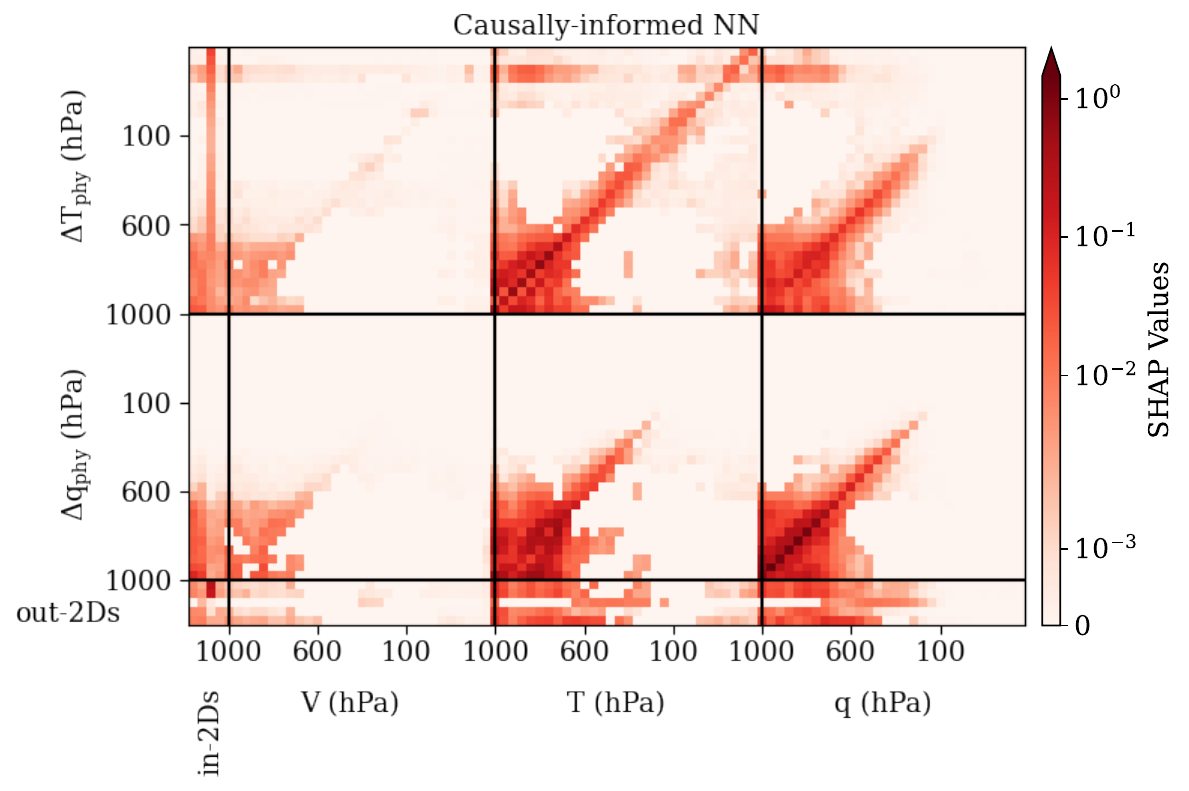}
        \label{fig:si_shap_causal}
    \end{minipage}
    
    \begin{minipage}[b]{0.66\textwidth}
        \centering
        \subcaption{}
        \vspace{-4mm}
        \includegraphics[width=\textwidth]{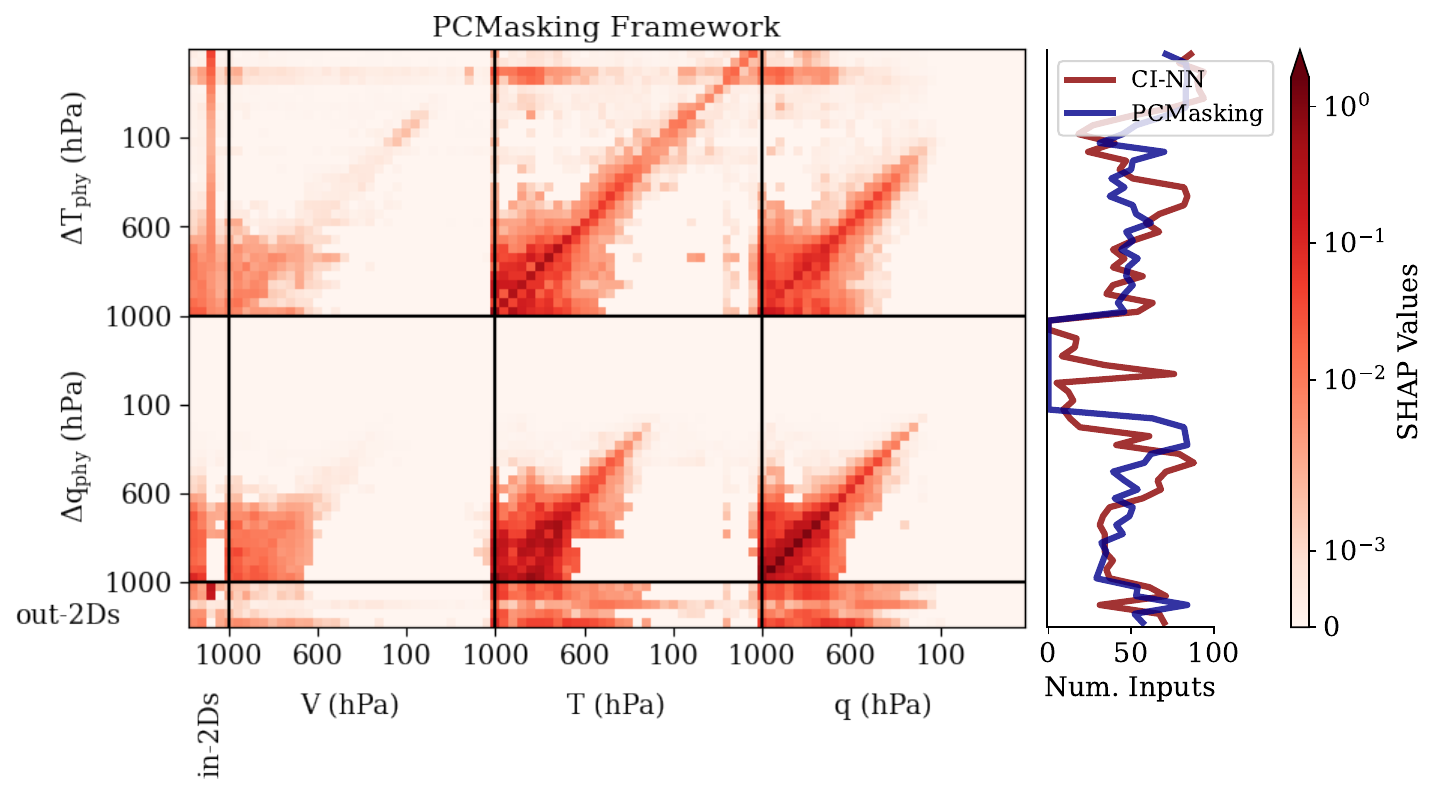}
        \label{fig:si_shap_mask_net}
    \end{minipage}
    
    \caption{Same as Fig. 4 but including 2D variables. 
    Mean absolute SHAP values computed from 1000 data samples for standard feed-forward neural networks (NNs) (\subref{fig:si_shap_single}), causally-informed NNs \citep{iglesias-suarez2024} (CI-NN) (\subref{fig:si_shap_causal}) and PCMasking framework NNs (\subref{fig:si_shap_mask_net}). 
    The right plot in Fig. (\subref{fig:si_shap_mask_net}) shows the total number of inputs for CI-NN and the PCMasking framework. 
    }
    \label{fig:si_shap_values}
\end{figure*}

\begin{figure*}[tb]
    \centering
    \includegraphics[width=0.7\linewidth]{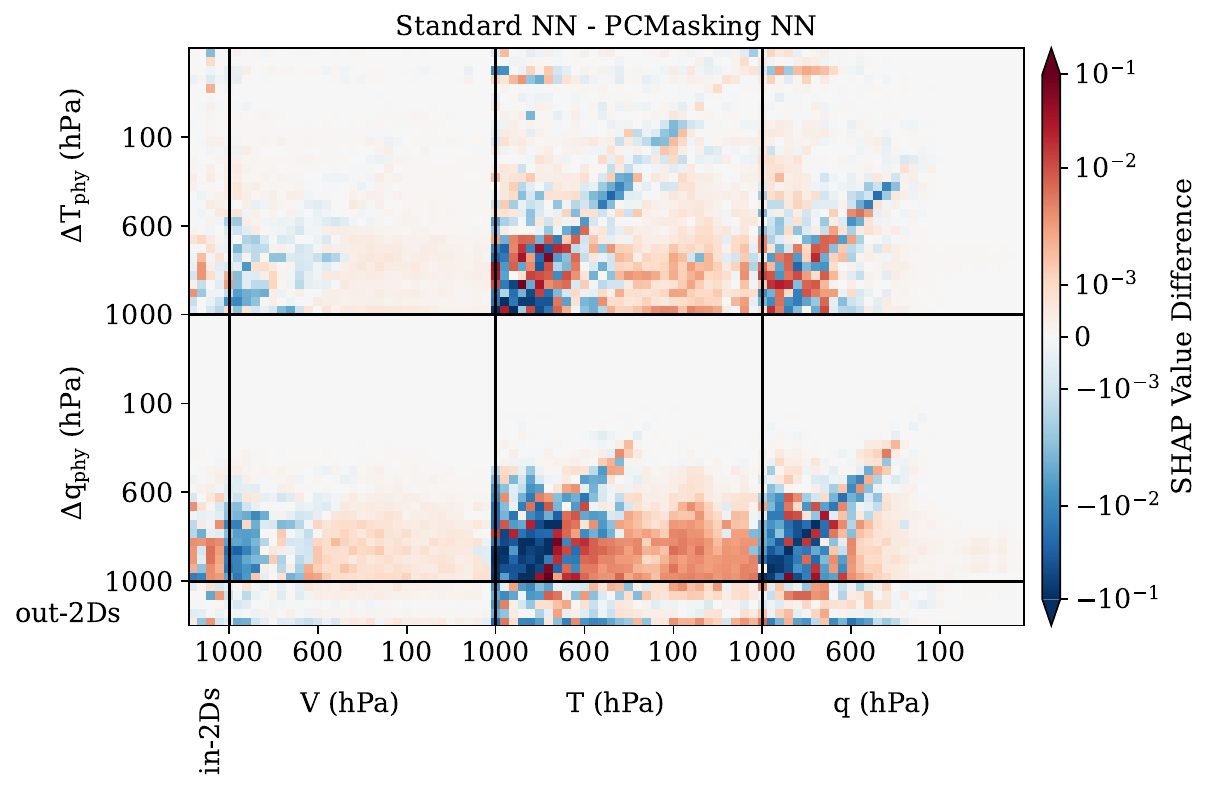}
    \caption{
    Same as Fig. 5 but including 2D variables. 
    SHAP value difference between mean absolute SHAP values for standard NNs and PCMasking framework networks. 
    SHAP values were computed from 1000 samples.
    Red values indicate stronger input-output connections in the standard NNs, while blue values indicate stronger connections in the PCMasking framework networks.  
    }
    \label{fig:si_shap_diff}
\end{figure*}

\begin{figure*}[ht]
    \centering
    \includegraphics[width=0.7\linewidth]{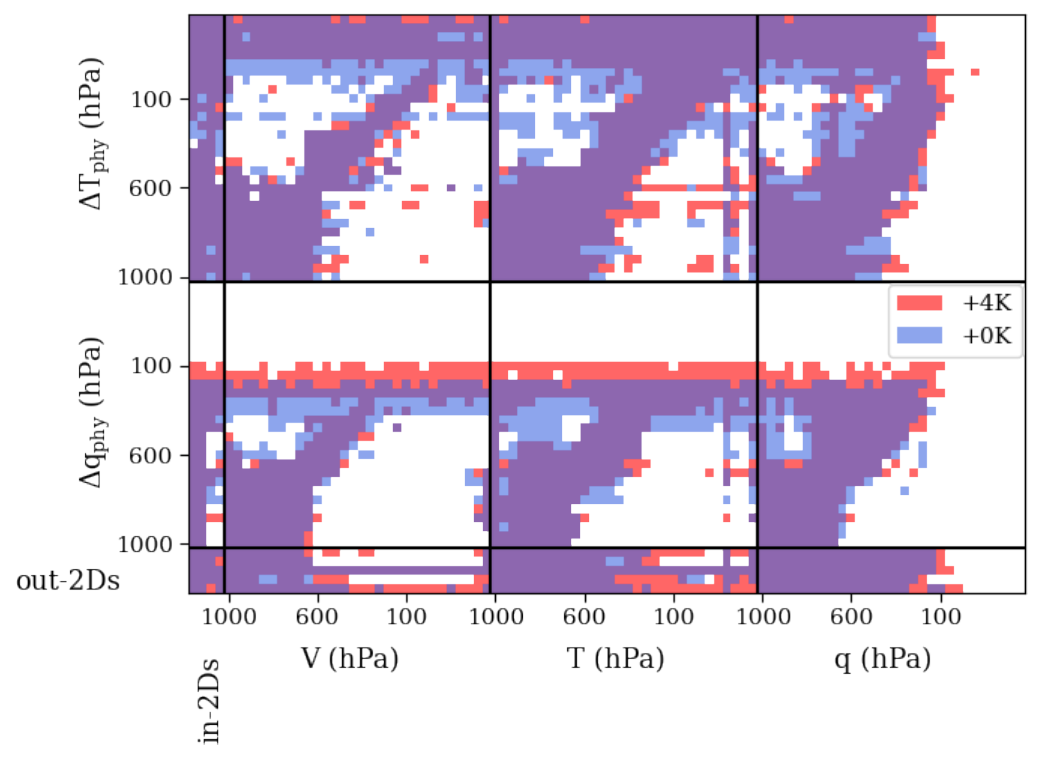}
    \caption{Same as Fig. 6 but including 2D variables. 
    Physical drivers for moistening and heating tendencies found with the PCMasking framework. Selected inputs for the 0~K reference climate are shown in blue and for the +4~K climate in red. 
    Violet areas indicate where the physical drivers overlap. 
    }
    \label{fig:si_inputs_plus4k}
    \vspace{5mm}
\end{figure*}

\begin{figure*}[ht]
    \centering
    \includegraphics[width=0.7\linewidth]{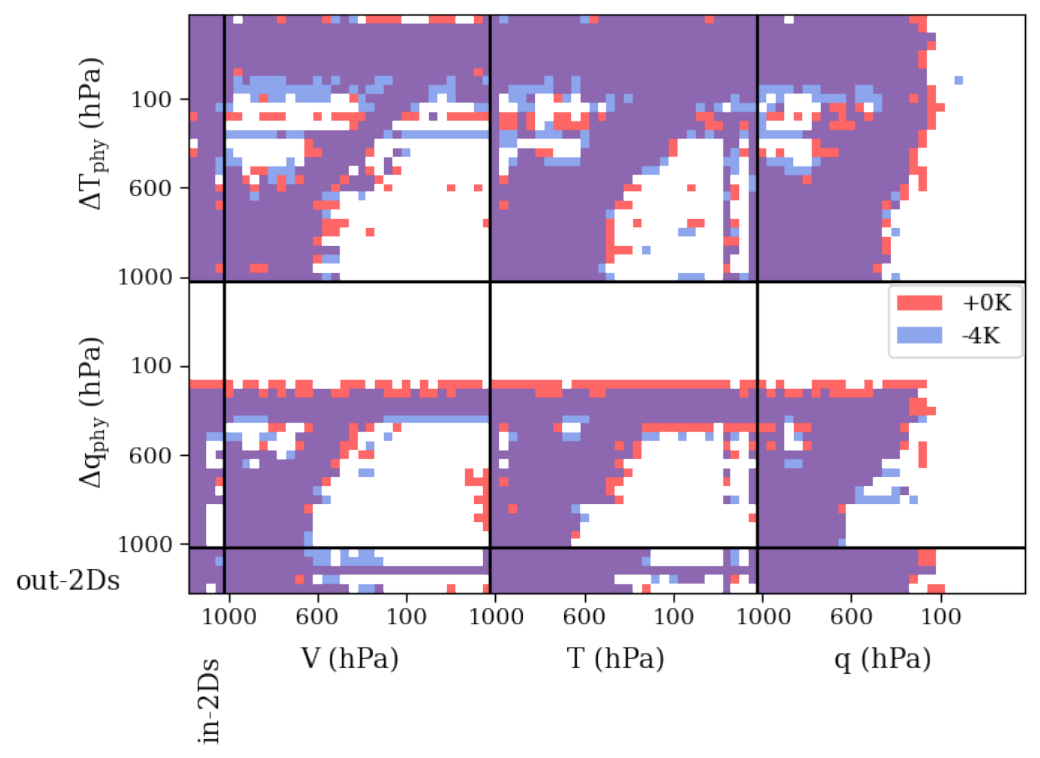}
    \caption{Same as Fig. 6 but for -4~K climate and including 2D variables. 
    Physical drivers for moistening and heating tendencies found with the PCMasking framework. Selected inputs for the 0~K reference climate are shown in red and for the -4~K climate in blue. 
    Violet areas indicate where the physical drivers overlap. 
    }
    \label{fig:si_inputs_minus4k}
    \vspace{5mm}
\end{figure*}



\end{document}